\pdfoutput=1

\documentclass[11pt]{article}

\usepackage[]{acl} 

\usepackage{times}
\usepackage{latexsym}
\usepackage{svg}

\usepackage{graphicx}
\usepackage{subcaption}

\usepackage{multirow}

\usepackage[T1]{fontenc}

\usepackage[utf8]{inputenc}

\usepackage{microtype}

\usepackage{inconsolata}

\title{Adaptive Cross-lingual Text Classification through In-Context One-Shot Demonstrations}

\author{
 \textbf{Emilio Villa-Cueva\textsuperscript{1,2}},
 \textbf{A. Pastor López-Monroy\textsuperscript{1}},
\\
\textbf{Fernando Sánchez-Vega\textsuperscript{1,4}},
 \textbf{Thamar Solorio\textsuperscript{2,3}}
\\
\\
 \textsuperscript{1}Mathematics Research Center (CIMAT), Gto, Mexico,
 \\
 \textsuperscript{2}Mohamed bin Zayed University of Artificial Intelligence, UAE,
 \\
 \textsuperscript{3}University of Houston, USA,
\\
 \textsuperscript{4}Consejo Nacional de Humanidades, Ciencias y Tecnologías, México
\\
}

\begin{document}
\maketitle
\begin{abstract}
Zero-Shot Cross-lingual Transfer (ZS-XLT) utilizes a model trained in a source language to make predictions in another language, often with a performance loss. To alleviate this, additional improvements can be achieved through subsequent adaptation using examples in the target language. In this paper, we exploit In-Context Tuning (ICT) for One-Shot Cross-lingual transfer in the classification task by introducing In-Context Cross-lingual Transfer (IC-XLT). The novel concept involves training a model to learn from context examples and subsequently adapting it during inference to a target language by prepending a One-Shot context demonstration in that language. Our results show that IC-XLT successfully leverages target-language examples to improve the cross-lingual capabilities of the evaluated mT5 model, outperforming prompt-based models in the Zero and Few-shot scenarios adapted through fine-tuning. Moreover, we show that when source-language data is limited, the fine-tuning framework employed for IC-XLT performs comparably to prompt-based fine-tuning with significantly more training data in the source language. 
\end{abstract}

\section{Introduction}
The recent progress in the development of multilingual Language Models (LMs) has allowed for effective cross-lingual transfer (XLT) with minimal need for architectural modifications \cite{pires, mt5}. By simply training a multilingual model in a language with abundant resources its acquired knowledge can be extended to target languages, in either Zero-Shot or Few-Shot scenarios. Cross-lingual transfer is a significant topic as it addresses the prevalent challenge of data scarcity in languages other than widely resourced ones, such as English \cite{joshi-etal-2020-state}. The ability to leverage the extensive linguistic resources available in high-resource languages for languages with limited training data enables the deployment of truly inclusive NLP systems. 

Zero-Shot Cross-lingual Transfer (ZS-XLT) involves transferring a model trained in a source language to a target language without any demonstration of target-language examples \cite{chen-etal-2021-zero,pires}. This approach is highly modular, as it requires no adaptations specific to the target language. On the other hand, Few-Shot Cross-lingual Transfer (FS-XLT) enhances target-language accuracy by further fine-tuning a model using labeled target data \cite{lauscher,zhao,schmidt}. However, this method faces limitations when the available target-language data is limited, especially for higher-level tasks, leading to negligible enhancements or even detrimental effects on performance \cite{lauscher,zhao}. Furthermore, this approach incurs higher computational costs due to the fine-tuning step and diminishes the modularity that characterizes the Zero-Shot method.

Our perspective is that adapting to a target language should prioritize resource efficiency and modularity, where we can seamlessly deploy a single model trained in English (or another source language) across different languages without any fine-tuning. In this work, we aim to improve this aspect for text classification --a high-level task-- by leveraging the language-specific abilities of a multilingual model by prepending a One-Shot text-label target-language demonstration to the input text to predict the correct label. Specifically, we propose In-Context Cross-lingual transfer (IC-XLT), a simple yet effective method for One-Shot Cross-Lingual Transfer in Text Classification. 

This novel approach employs In-Context Tuning (ICT) \cite{chen-ict} to train an encoder-decoder model in the source language tasking it to predict input texts with information derived from context demonstrations. ICT is a meta-learning strategy that optimizes a model's ability to learn from in-context examples, originally designed for facilitating swift adaptation to new tasks by prepending target-task in-context demonstrations to the input during the adaptation process. To the best of our knowledge, this is the first study of ICT application in the context of cross-lingual transfer. 

The proposed method is composed of a fine-tuning and an adaptation stage. First, we fine-tune on the source language through ICT, where the model is trained for the classification task and also to learn from context demonstrations. Then, we adapt to the target language at inference time by prepending a One-Shot\footnote{One-Shot per label} demonstration in that language to the input. This method is modular and cost-effective at the adaptation stage as it does not require any gradient update.

We evaluate IC-XLT  on two multilingual text classification datasets, spanning 5 and 13 target languages, with English as the source language. We consider two distinct settings. First, we assume access to the entire source-language training dataset. For the second setting, we deliberately constrain the amount of source training data available. This limitation aims to gauge the robustness of the proposed approach in scenarios where the availability of source data is restricted. We hypothesize that leveraging context information may prove particularly beneficial in tasks where source data is limited.

The contributions of this work are the following: 

\begin{enumerate}
    \item \textbf{IC-XLT as an effective strategy for One-Shot Cross-lingual transfer:} By measuring the reduction in the transfer gap of IC-XLT against standard approaches, and the performance improvement after introducing target-language examples, we present empirical evidence that training a model in a source language with In-Context Tuning allows it to leverage a One-Shot demostration through In-Context Learning to adapt to a target language. This results in a One-Shot XLT approach that adapts to a target language at inference without requiring gradient updates.
    
    \item \textbf{ICT improves mT5 fine-tuning when resources are limited.} We observe that for the evaluated tasks, ICT training yields better performance compared to traditional fine-tuning when (source language) training data consists of only few-shots per label. In particular IC-XLT models trained on this scenario (1) benefit from this behavior at the adaptation and (2) leverage target-language in-context examples, achieving comparable performance to Prompt Tuning transfer methods with significantly less source-language data.
\end{enumerate}

\section{Related work}
\label{sec:related_work}

\subsection{Zero and Few-Shot Cross-lingual Transfer}

Multilingual transformers, such as mBERT \cite{bert}, XLMR \cite{xlmr}, and mT5 \cite{mt5}, have showcased notable ability in Zero-Shot Cross-lingual Transfer (ZS-XLT) \cite{pires}. In this paradigm, these models are trained using abundant data in a source language and subsequently undergo evaluation in a target language without exposure to any training data in that specific language. However, this methodology is susceptible to significant performance variance \cite{keung}, and the transfer performance gap is contingent upon the linguistic proximity between the source and target languages \cite{pires}.

Furthermore, recent studies indicate that incorporating a small number of annotated examples in the target language can mitigate the performance gap between the source and target languages \cite{lauscher,zhao,schmidt}. This methodology, termed Few-Shot Cross-Lingual Transfer (FS-XLT), involves first fine-tuning a model on an extensive source dataset (as in ZS-XLT), and then subjecting it to a second fine-tuning on the reduced target-language data, facilitating its adaptation to this target language. This approach yields a noticeable improvement in performance at a relatively low labeling cost across various NLP tasks \cite{lauscher}.

However, empirical evidence \cite{lauscher,zhao, schmidt} indicates that FS-XLT yields the most significant benefits for \textit{low-level} tasks, such as Named Entity Recognition (NER) or Part-of-Speech (POS) tagging. When applied to \textit{high-level} tasks, like Natural Language Inference (NLI) or text classification, the performance only shows improvement with a substantial number of examples from the target language. When adapted with small datasets in the target language (<100 samples), FS-XLT tends to offer minimal performance gains or may even lead to a decline in model efficacy.

Additionally, according to \citet{schmidt}, sequential FS-XLT can also exhibit unreliability in the Few-Shot scenario due to considerable variance in performance at different checkpoints during training. To address this issue, they propose jointly training the model using both source and target data in the adaptation stage of the process, which improves stability in the Few-Shot setting. This \textit{fine-tuned} FS-XLT approach, however, has two notable drawbacks. Firstly, it lacks modularity, as the models are trained specifically for the selected target language during the adaptation stage. Secondly, there is a substantial increase in computational cost compared to Zero-Shot Cross-lingual Transfer due to the adaptation fine-tuning, whose cost scales with the size of the base model. 

Moreover, existing methods predominantly address the XLT task under the assumption of abundant data in the source language. Although this is a fair assumption for many cases, as in general it is much more likely to find labeled datasets in high resource languages, there are scenarios where the source domain itself is limited. Instances of this include domain-specific tasks with a scarcity of annotated samples or tasks related to rapidly emerging trends and language patterns originated from social media, where due to its emerging nature, there is no labeled data available. In such cases, it might be more feasible to secure annotations for high-resource languages, which can then be transferred to other languages.

Given these considerations, we believe it is pertinent to investigate how the XLT performance scales as the quantity of available source data is systematically reduced. The intuition behind this is that the introduction of target-language shots may alleviate the performance decrease associated with a reducing source training data.

\subsection{In-Context Learning and Language Models}

LMs have demonstrated an aptitude for learning from a small number of demonstrations through a method known as In-Context Learning (ICL) \cite{brown_gpt3}, where the model is tasked with predicting an input prepended with labeled examples. Particularly, \cite{winata-mling} observed that it is possible to achieve satisfactory performance in a cross-lingual setting when evaluating an mT5 model with a target-language input prefixed with labeled English demonstrations. This Zero-Shot approach, although efficient, can be sub-optimal as it does not take full advantage of resources in the source language due to the lack of fine-tuning.

Recent findings indicate that transformers \cite{vaswani_transformers} can perform \textit{model selection} on functions encountered during pre-training through in-context demonstrations. Yet, they still find it challenging in generalizing effectively to out-of-distribution classes, as highlighted by \citet{yadlowsky}. Given that most pre-trained LMs have not been explicitly trained for ICL, they might exhibit sub-optimal behavior when presented with Few-Shot demonstrations. In response to this challenge,  \citet{chen-ict} introduced In-Context Tuning (ICT), a meta-learning approach designed to train a model to effectively learn from in-context demonstrations\footnote{Also, ICT consistently improves performance of ICL and is less sensitive to the shot selection when compared to raw, pre-trained LMs. \cite{chen-ict}}. ICT meta-trains a language model across a range of tasks, enhancing its ability to swiftly adapt to new tasks through ICL. 

Still, In-Context Tuning has not yet been implemented for language transfer, as opposed to task transfer. We hypothesize that fine-tuning a multilingual model concurrently for learning from input context and the downstream task can leverage multilingual knowledge acquired during pretraining. This, we anticipate, will result in enhanced classification performance in a target language when provided with examples in that language. Therefore, in this study we showcase the efficacy of this idea for One-Shot Cross-lingual Transfer, particularly, for adapting to a target language through a One-Shot demonstration in-context. This adaptation method proves effective in improving the text classification performance by better leveraging target-language examples compared to the the fine-tuned FS-XLT. Moreover, we delve into the advantages of employing this approach in scenarios where source task data is not abundant. 

\section{Our proposed approach: In-Context Cross-Lingual Transfer}
\label{sec:ict}
Our method aims to simultaneously train a pretrained multilingual encoder-decoder model for (1) a downstream text classification task, and (2) learning from context demonstrations. Then, we expect it to be able to generate predictions in a target language by including context demonstrations in this language. Therefore, we leverage ICT fine-tuning to transfer between languages at inference. As described above, our proposed procedure, called In-Context Cross-lingual Transfer (IC-XLT), is comprised of two stages:

\paragraph{In-Context Tuning}
During the meta-training stage, we fine-tune the base multilingual model for a specific task using data from the source language. Let the set of pairs \(D^{src} = \{(x^{src}_1, y^{src}_1), \ldots, (x^{src}_{|D|}, y^{src}_{|D|})\}\) represent the source-language training dataset. The objective is to train the model to predict the label \(y_i^{src}\) for a given text \(x_i^{src}\) with the following input\(\Rightarrow\)output format:
$$X^{src}, x_i^{src} \Rightarrow y_i^{src}$$
Here, \(X^{src} = \left((x_{j_1},y_{j_1}),\ldots,(x_{j_M},y_{j_M} ) \right)\) is a random sequence of \(M\) text-label pairs randomly sampled from \(D^{src}\) without replacement, which excludes the pair \((x_i^{src}, y_i^{src})\). In simpler terms, \(M\) is the number of source-language demonstrations prepended to each input during fine-tuning, not the number of \(K_{tgt}\)-shot examples prepended during inference.

\paragraph{In-Context Learning}

At inference, we adapt to a target language by prepending the samples from the target language training dataset \(\widetilde D^{tgt} = \{(\widetilde 
 x^{tgt}_1, \widetilde  y^{tgt}_1), \ldots, (\widetilde x^{tgt}_{NK_{tgt}}, \widetilde y^{tgt}_{NK_{tgt}})\}\) to each entry \(x_i^{tgt}\) of the test set to predict \(y_i^{tgt}\). Consequently, the input format mirrors the structure observed in the ICT stage:

$$\widetilde X^{tgt}, x_i^{tgt}  \Rightarrow y_i^{tgt} $$

Where \(N\) is the number of classes and the sequence \(\widetilde X^{tgt}\) is a concatenation of \(\widetilde D^{tgt}\) entries comprising the \(K_{tgt}\)-shot samples per class, prepended to each \(x_i^{tgt}\) entry at the inference stage.

The intuitive idea for this approach is that, after the meta-training stage, we expect the model to understand both the classification task and the contextual relationships relevant to it. During the adaptation stage, the model leverages its multilingual pretraining to interpret context examples in the target language. Note that the adaptation to the target language in this context is gradient free, as it occurs during inference, and thus no model weights are updated. 

\section{Experimental Methodology}

In this section, we outline the methodology employed to evaluate the proposed approach. We assess IC-XLT effectiveness in adapting to a target language for the classification task and compare its performance in cross-lingual transfer under (1) full training data on the source language and (2) various source-language data budgets. We conduct these limited data experiments to assess how much IC-XLT improves over a traditional fine-tuning method by leveraging the One-Shot demonstration.

\subsection{Data and Evaluation Metrics}

We conduct evaluations on two mutlilingual text classification datasets. The first dataset is Aspect Category Detection (ACD) on Restaurant Reviews \cite{pontiki-etal-2016-semeval}, a multi-label dataset comprising 12 classes representing different aspects mentioned in reviews. The second dataset is Domain Classification on assistant utterances from the MASSIVE dataset \cite{massive}, a single-label classification dataset with 18 possible domain classes. The main difference between these datasets is that MASSIVE assigns only one label per entry, whereas ACD allows for multiple labels per entry, presenting a more challenging task. The datasets were chosen for their larger number of labels and their availability in multiple languages with shared labels (See Appendix \ref{sec:appendix_datasets} for further details).   

We select \(F_1\) micro as our evaluation metric, following \citet{pontiki-etal-2016-semeval}. For both datasets, our model is trained in English as the source language, and its performance is evaluated across 5 target languages: \textit{Dutch, Turkish, Russian, French,} and \textit{Spanish} for ACD, and 13 target languages for MASSIVE: \textit{French, Spanish, Turkish, Russian, Thai, Japanese, Indonesian, Icelandic, Amharic, Arabic, Azeri, Swahili} and \textit{Urdu}

To evaluate the performance of our proposed In-Context Cross-Lingual Transfer (IC-XLT) approach in a resource-constrained scenario with limited source-language data, we construct synthetically reduced datasets by sampling subsets of the training datasets following various \(K\)-shot configurations, specifically \(K_{src} \in \{8,16,32,64\}\). The objective of these evaluations is to assess IC-XLT's ability to leverage target-language demonstrations for enhancing performance in situations where the source-language task has limited resources.

Regarding the shot selection, our \(K\)-shot approach selects \(K\) examples per class. Considering the class imbalance and the multi-label nature of the ACD dataset, the total number of examples will be in the range \(\left[K,K \times N\right]\). For a detailed explanation on this refer to Appendix \ref{sec:appendix_shotselection}

\subsection{Experimental Setting}\label{sec:exp_setting}
As our multilingual base model, we utilize mT5-large \cite{mt5}, an encoder-decoder model pre-trained on a diverse corpus encompassing over 100 languages. We employ LoRA \cite{lora} for fine-tuning the model on the source-language data with full training data and varying numbers of shots \(K_{src}\). During the inference stage, label predictions are generated through text generation, which facilitates multi-label inference. We adopt a greedy decoding strategy as implemented in \citet{wolf_huggingface}.

In this work, we set \(K_{tgt} = 1\), using only a One-Shot demonstration for the proposed IC-XLT approach\footnote{We provide IC-XLT implementation in the following repository: \url{https://github.com/villacu/ic_xlt}}. We train the ICT models in the source language with different number of context examples, specifically \(M = 10\) and \(M = 20\). All models are trained on an NVIDIA Titan RTX GPU, the hyperparameter selection is discussed in Appendix \ref{sec:hpp_selection_ft}.

For the experiments with reduced source-language data, we conduct evaluations using two seeds for each of the following: the fine-tuning process, \(K_{src}\) shot selection, and \(K_{tgt}\) shot selection. Since Zero-Shot approaches do not require selecting target shots, we run a total of 4 and 8 runs for Zero-Shot and One-Shot respectively, using seeds within \(\{1,2\}\). For the models trained with full source-language data, we trained 5 models with seeds for the fine-tuning process within \(\{1,...,5\}\) and selected the best 3 in the English validation set. At adaptation, we employ two seeds for \(K_{tgt}\) shot selection in \(\{1,2\}\).

\subsection{Baselines}
\label{sec:baselines}
We benchmark our proposed approach against the following baseline methods:
\paragraph{(1S) One-shot Prediction}
Leveraging mT5's pretraining objective, we task the model with predicting the missing span corresponding to the correct label given an input text prepended with a One-Shot demonstration. This is a traditional In-Context Learning approach where we expect the model to deduce label meanings from the examples without undergoing source-language fine-tuning, similar to the idea introduced in \citet{winata-mling}, serving as the lower bound when \(K_{src}=0\).

\paragraph{(ZS-XLT) Zero Shot XLT}
The standard Zero-Shot (\(K_{tgt}=0\)) Cross-lingual Transfer approach, where the model is initially trained on a source language, and subsequent inference is conducted on the target language without any additional tuning. In this case, we train the mT5 model through \textit{prompt-based} fine-tuning (PFT), with the input-output form:
\[x_i \Rightarrow y_i\]
Hence, training is performed at the source and inference at target languages. 

\paragraph{(1S-XLT\(^{\nabla}\) and 8S-XLT\(^{\nabla}\)) 1-Shot and 8-Shot XLT}

Using the same training scheme (PFT), we continue fine-tuning on the checkpoints from ZS-XLT, training with \(K_{tgt}=1\) and  \(K_{tgt}=8\) \textit{shots per label} in the target language. This approach is the standard \textit{gradient-based} approach for adapting to a target language in Few-Shot Cross-Lingual Transfer \cite{lauscher}. For this baseline, the target-language \textit{shots} are not prepended as in IC-XLT and 1S, but used to further fine-tune the model following the same input-output as ZS-XLT.


\paragraph{(1S-XLT\(^{\nabla}_{macro}\)) \textit{macro-averaging}}

In this baseline, we build upon the methodology established in 1S-XLT\(^\nabla\) with \(K_{tgt}=1\) by adopting a strategy that incorporates fine-tuning on both source and target-language data for adaptation to the target language. This method follows the approach described by \citet{schmidt}, where the loss for each batch is calculated using a weighted combination of source and target-language losses:

\[L = \beta L_{src} + (1-\beta)L_{tgt}\]

Here, \(L_{src}\) and \(L_{tgt}\) represent the source and target-language losses, respectively. We select a value of \(\beta = 0.5\) following the original implementation.

\paragraph{(IC-XLT\(_{SRC}\)) IC-XLT with source-language context}
We use the same models trained for IC-XLT, however, in this method In-Context examples are drawn from the training set of the source language (English). In essence, this can be considered a Zero-Shot baseline since no target language (or \textit{unseen} source language) is involved for adaptation. Through this baseline we aim to evaluate the relevance of the \textit{target} language One-Shot samples at the adaptation stage, assessing whether they are necessary for successful transfer to that target language.

\section{Results and analysis}

\paragraph{IC-XLT performance at Cross-lingual transfer}

\begin{table*}[]
\centering
\resizebox{2\columnwidth}{!}{%
\begin{tabular}{l|l|l|lllllllllllll}
\hline
Method ($K_{tgt}$) & ENG(SRC) & Target avg & AMH & AZE & ISL & SWA & URD & ARA & IND & THA & TUR & FRA & JAP & RUS & SPA \\
\hline
1S (1) & 38.54 & 30.48 & 24.26 & 30.38 & 31.19 & 32.78 & 29.48 & 31.15 & 32.26 & 30.83 & 33.24 & 33.34 & 29.69 & 25.80 & 31.90\\
ZS-XLT (0) & 90.06 & 76.12 & 62.53 & 71.63 & 73.77 & 65.75 & 67.72 & 70.98 & 84.23 & 79.88 & 77.7 & 84.87 & 83.39 & 83.98 & 83.09 \\
1S-XLT$^{\nabla}$ (1) & 90.08 & 76.21 & 62.78 & 71.63 & 73.9 & 65.85 & 68.34 & 71.11 & 84.17 & 79.95 & 77.68 & 84.86 & 83.45 & 83.89 & 83.15 \\
8S-XLT$^{\nabla}$ (8) & 89.87 & 76.78 & 65.38 & 72.13 & 74.73 & 66.78 & 69.24 & 71.45 & 84.08 & 80.46 & 78.02 & 85.16 & 83.25 & 84.17 & 83.29 \\
1S-XLT$_{macro}^{\nabla}$ (1) & 89.93 & 75.8 & 62.53 & 71.05 & 73.48 & 65.42 & 67.9 & 70.49 & 83.8 & 79.33 & 77.17 & 84.71 & 82.87 & 83.78 & 82.86 \\
\hline
IC-XLT$_{SRC}^{M=10}$ (0) & 89.41 & 74.39 & 63.39 & 70.82 & 69.22 & 56.06 & 67.47 & 71.62 & 81.04 & 79.02 & 76.65 & 82.81 & 82.99 & 84.41 & 81.53 \\
IC-XLT$_{SRC}^{M=20}$ (0) & 89.46 & 74.02 & 61.27 & 71.49 & 68.83 & 53.76 & 68.26 & 70.77 & 81.6 & 78.91 & 76.36 & 83.13 & 82.83 & 83.76 & 81.23 \\
IC-XLT$^{M=10}$ (1) & 89.41 & \textbf{81.24} &\textbf{70.68} & \textbf{81.73} & \textbf{81.07} & \textbf{78.23} & \textbf{76.34} & \textbf{77.28} & \textbf{85.6} & 80.98 & \textbf{83.63} & \textbf{85.53} & \textbf{84.68} &\textbf{86.18} & \textbf{84.18} \\
IC-XLT$^{M=20}$ (1) & 89.46 & 80.26 & 68.09 & 81.37 & 80.32 & 75.54 & 76.12 & 76.43 & 85.46 & \textbf{81.02 }& 82.25 & 84.92 & 83.77 & 84.39 & 83.65\\
\hline
\end{tabular}
}
\caption{
Average \(F_1\) micro in the MASSIVE Domain Detection task, trained with full data in English. The number in parenthesis is the amount of samples per label in the target language used for the adaptation process (\(K_{tgt}\)). Standard deviations over different seeds per each language are shown in Appendix \ref{sec:appx_allexps}.
}
\label{tab:massive_full}
\end{table*}

\begin{table}[]
\centering
\resizebox{1.0\columnwidth}{!}{%
\begin{tabular}{l|l|l|lllll}
\hline
Method ($K_{tgt}$) & ENG(SRC) & Target avg & FRA & NLD & RUS & SPA & TUR \\
\hline
1S (1) & 37.09 & 28.42 & 31.78 & 20.33 & 34.38 & 34.86 & 20.76 \\
ZS-XLT (0) & 79.14 & 70.15 & 67.96 & 69.24 & 73.6 & 70.01 & 69.95 \\
1S-XLT$^{\nabla}$ (1) & 79.16 & 70.39 & 67.98 & 69.56 & 73.44 & 70.23 & 70.72 \\
8S-XLT$^{\nabla}$ (8) & 78.78 & 70.26 & 67.87 & 68.81 & 72.32 & 69.84 & 72.47 \\
1S-XLT$_{macro}^{\nabla}$ (1) & 79.16 & 70.41 & 68.03 & 69.59 & 73.3 & 70.25 & 70.88 \\
\hline
IC-XLT$_{SRC}^{M=10}$  (0) & 81.48 & 73.83 & 74.06 & 72.54 & 77.59 & 73.7 & 71.27 \\
IC-XLT$_{SRC}^{M=20}$  (0) & 81.76 & 73.11 & 73.49 & 71.89 & 77.52 & 73.14 & 69.51 \\
IC-XLT$^{M=10}$ (1) & 81.48 & \textbf{75.25} & \textbf{74.07} & \textbf{73.34} & \textbf{78.07} & \textbf{75.20} & 75.59 \\
IC-XLT$^{M=20}$ (1) & 81.76 & 75.05 & 73.5 & 73.00 & 78.01 & 74.46 & \textbf{76.26} \\
\hline
\end{tabular}
}
\caption{
Average \(F_1\) micro in the Aspect Category Detection dataset, trained with full data in English, the source language. Standard deviations over different seeds per each language are shown in Appendix \ref{sec:appx_allexps}.
}
\label{tab:acd_full}
\end{table}

For the first experiment, we compare our proposed approach, IC-XLT, to the baselines detailed in Section \ref{sec:baselines} using the full training set in the source language. For Aspect Category Detection, we observed a general trend where mT5 models trained with In-Context Tuning, which employs the input-output setting \(\widetilde X, x_i  \Rightarrow y_i\), consistently outperformed models subjected to prompt-based fine-tuning with \(x_i \Rightarrow y_i\) under the same training regimes, despite both models being trained for an equivalent number of steps and exact same data instances. We hypothesize that this superior performance may be attributed to the fact that the ICT-trained models \textit{see} \(M\) randomly ordered input-output examples at each instance, even though they are tasked with predicting only \(x_i\). 

We present the \(F_1\) micro scores across five different languages on ACD and 13 languages on MASSIVE in Tables \ref{tab:massive_full} and \ref{tab:acd_full} respectively. The numbers in these tables represent average metrics calculated across various seeds, as detailed in Section \ref{sec:exp_setting}. The standard deviation for each method and language is provided in the complete tables located in Appendix \ref{sec:appx_allexps}. We observe that our proposed approach, In-Context Cross-Lingual Transfer (IC-XLT), effectively outperforms the baselines by a substantial margin in the evaluated datasets, greatly improving mT5 cross-lingual transfer performance. A crucial observation is that for both of the evaluated datasets there is a noticeable increase in performance from IC-XLT\(_{SRC}\) to IC-XLT. This means that the proposed approach adapts successfully, during inference time, to the target language by taking advantage of the One-Shot in context demonstration.

On the other hand, the 1S-XLT\(^{\nabla}\) approach did not improve over ZS-XLT by a considerable margin, which is consistent with previous results for high-level tasks \cite{lauscher,zhao}. 
 Even when increasing target-language samples to \(K_{tgt}=8\) (baseline 8S-XLT\(^{\nabla}\)) on this gradient-based baseline, we observe a performance decline in the ACD dataset for most languages except Turkish. The MASSIVE dataset shows improvements across nearly all target languages with the increased number of target shots on 8S-XLT\(^{\nabla}\), yet, these gains are modest compared to those achieved with IC-XLT using only \(K_{tgt}=1\), highlighting its superior efficiency in utilizing minimal target-language data for improved cross-lingual transfer.

The gradient-based adaptation that mixes target and source languages (1S-XLT\(^{\nabla}_{macro}\)) marginally outperforms 1S-XLT\(^{\nabla}\) in the ACD dataset. However, it does not improve, and even drops, performance on the MASSIVE dataset. This approach also requires more computational resources, due to the larger data volume when including source-language examples.

The observed variability in the effectiveness of gradient-based methods across datasets and languages may be related to differences between the datasets; ACD represents a more complex, non-parallel task with classes that are easier to confuse, unlike the parallel and simpler nature of MASSIVE. Nonetheless, IC-XLT consistently outperforms the baselines across languages and datasets, demonstrating its robustness and cost-effectiveness in terms of computational resources. We find that \(M=10\) (the number of in-context demonstrations during ICT training) performs slightly better than \(M=20\) in the experiments that employ the full training set.

\paragraph{Performance with limited source-language data}

\begin{figure*}[h]
\centering
  \begin{subfigure}[b]{0.45\textwidth}
    \includegraphics[width=\textwidth]{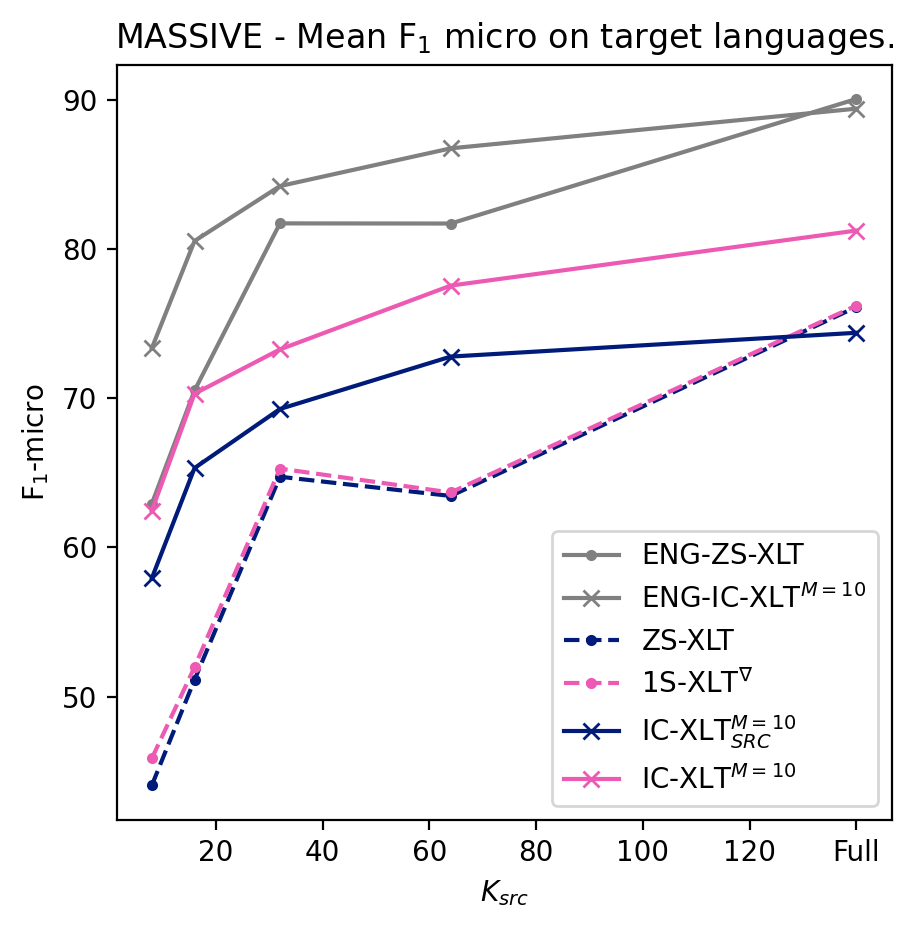}
    \caption{MASSIVE performance with different source data availability. IC-XLT trained with \(M=10\).}
    \label{fig:massive_plot}
  \end{subfigure}
  \quad
    \begin{subfigure}[b]{0.45\textwidth}
    \includegraphics[width=\textwidth]{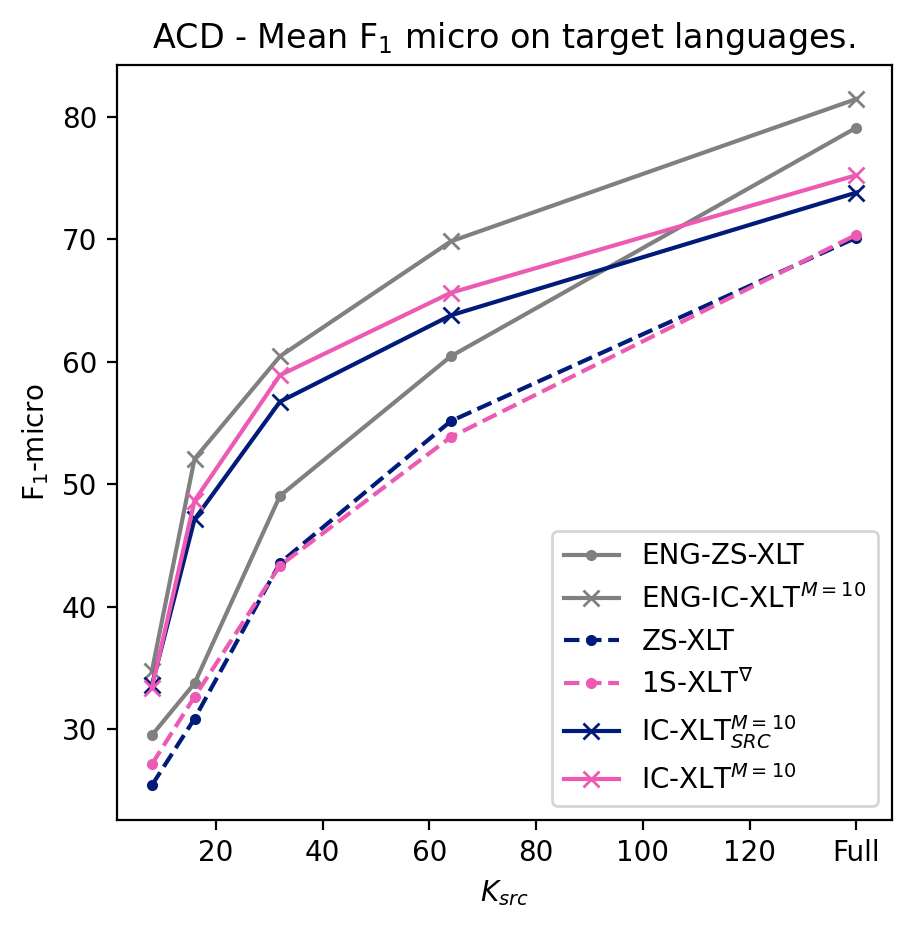}
    \caption{ACD performance with different source data availability. IC-XLT trained with \(M=10\).}
    \label{fig:acd_plot}
  \end{subfigure}

  \caption{Comparison of IC-XLT and 1S-XLT\(^{\nabla}\) performance at different source-language data budgets. Pink lines employ target-language examples for adaptation while blue lines do not. We can observe that, in general, the IC-XLT models yield better performance compared to ZS-XLT and 1S-XLT\(^\nabla\). This is especially notable at lower resource scenarios.}
  \label{fig:both_plots}
\end{figure*}

We conduct experiments to quantify the ability of IC-XLT to perform at scenarios with limited source-language resources. For this we evaluate ZS-XLT, 1S-XLT\(^{\nabla}\), IC-XLT\(_{SRC}\), and IC-XLT models trained with \(K_{src} \in \{8,16,32,64\}\). 
We noticed that models trained with the ICT framework generally perform better compared to PFT for low values of \(K_{src}\). In Figures \ref{fig:acd_plot} and \ref{fig:massive_plot}, we illustrate the average performance on the target languages at different source-language resource availability regimes. In both Figures, we can observe that IC-XLT makes better use of resources than prompt-based fine-tuning specially at smaller values for \(K_{src}\). Furthermore, the performance difference with the source language (English) is visibly smaller for IC-XLT, more discussion on this can be found below.

The \(F_1\)-micro averages for the target languages are shown in Tables \ref{tab:acd_results} and \ref{tab:massive_results} for ACD and MASSIVE, respectively. For all the language-specific performance metrics at different \(K_{src}\) budgets refer to Appendix \ref{sec:appx_allexps}. Results on these tables show that IC-XLT trained on limited data (\(K_{src}=64\)) achieves competitive or superior performance compared to baseline methods (ZS-XLT and 1S-XLT\(^{\nabla}\)) trained with full source datasets. 

Given that the target language adaptation occurs at inference time, the improvement over the Zero-Shot approach comes at no extra computational cost and at a minimal data cost. This allows to achieve good performance with limited computational and data resources.

For the experiments with limited source-language data, \(M=20\) achieves a better performance in MASSIVE. We believe that since ACD contains only 12 labels, in this scenario a context length of 20 will inevitably prepend more repeated context examples than the MASSIVE dataset\footnote{Which contains 18.} when training with limited data. This reduced variability may hurt the model's performance compared to \(M = 10\). 

\paragraph{Measuring the transfer gap with the source language.}
\label{sec:diffs}

By measuring the performance gap between the source language and the target language, we aim to quantify the contribution of the ICT training framework and target-language demonstrations for mitigating this gap. As we provide the model with target-language examples, we anticipate a \textit{smaller} decrease in performance from the source language when evaluating on a new language, compared to Zero-Shot approaches. We can measure this by computing the average transfer gap $\bar \Delta \%$, which is the average percentage decrease in performance relative to the evaluations on the test set in the source language (English). This is defined as:

\[\bar \Delta \% = 100 \times \mathbf{E}\left[P_{TL}/P_{SL} - 1\right]\]

\begin{figure*}
\centering
 \begin{subfigure}[b]{0.45\textwidth}
    \includegraphics[width=\textwidth]{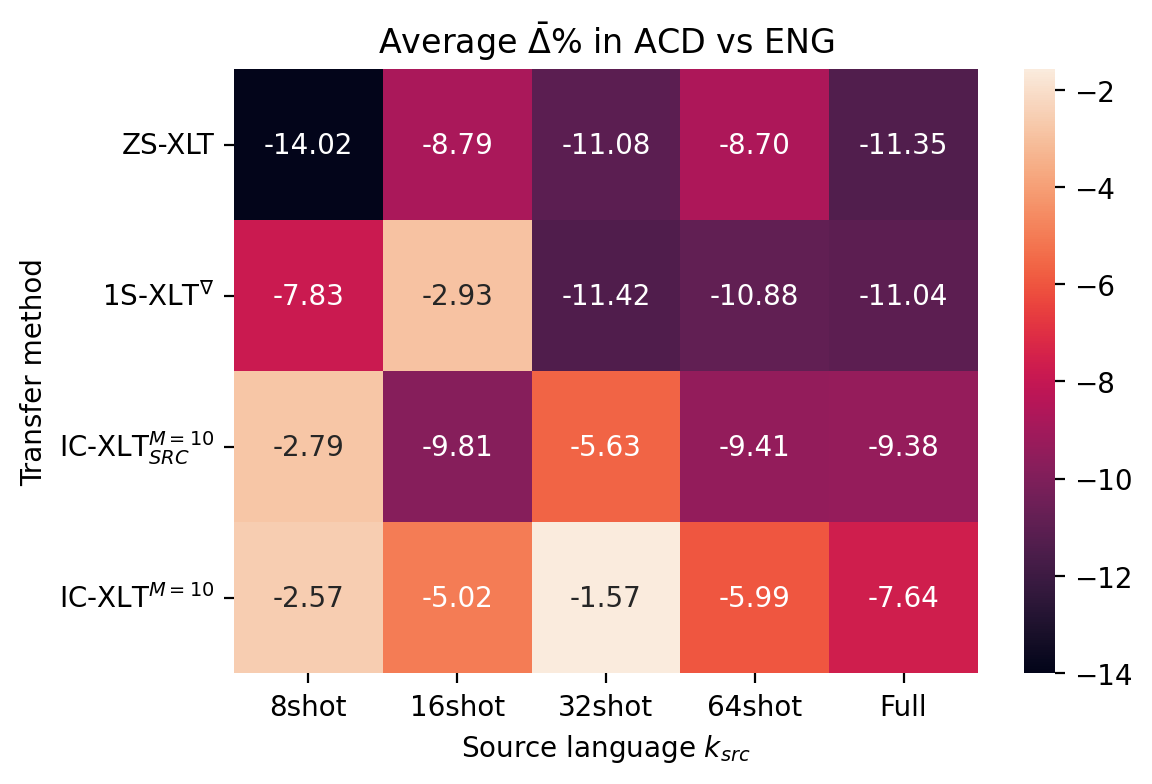}
    \label{fig:gap_acd}

  \end{subfigure}
  \quad
  \begin{subfigure}[b]{0.45\textwidth}
    \includegraphics[width=\textwidth]{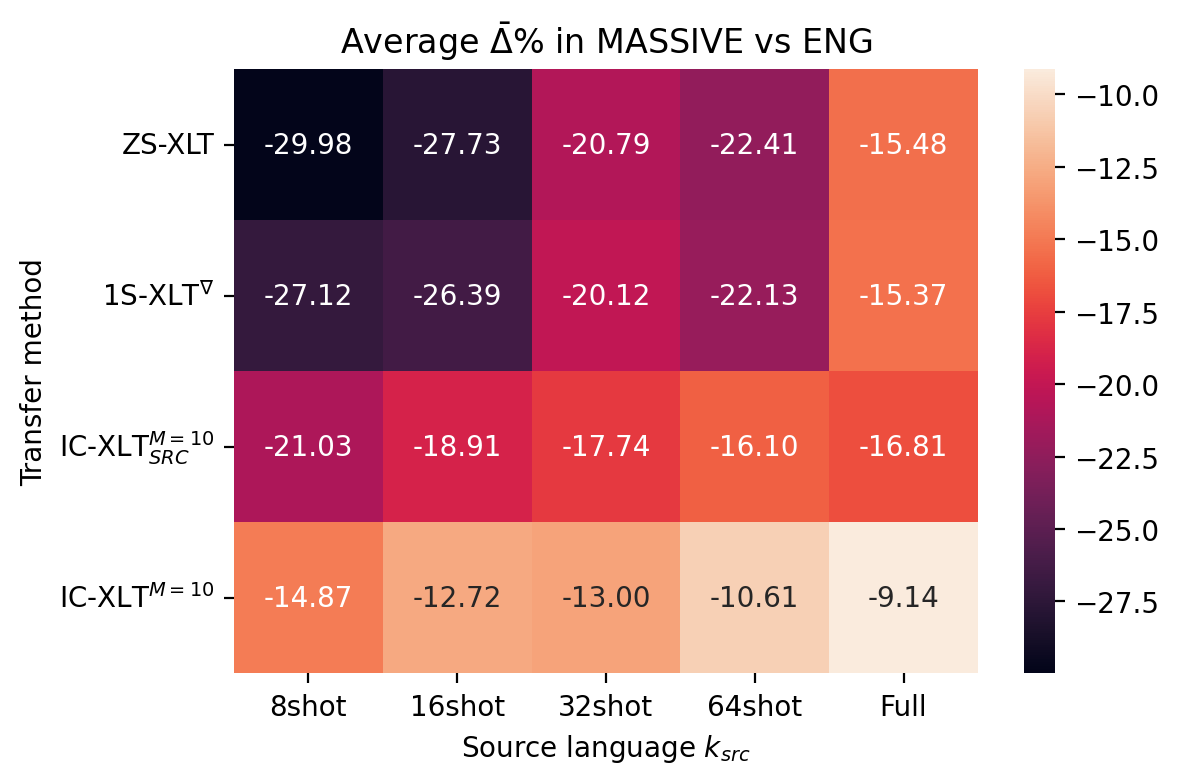}
    \label{fig:gap_massive}
    
  \end{subfigure}
  
  \caption{The average transfer gap \(\bar \Delta \%\) of IC-XLT, IC-XLT\(_{SRC}\), 1S-XLT\(^{\nabla}\) and ZS-XLT at different source-language data budgets. (IC-XLT \(M=10\)). We can observe that, for most cases, IC-XLT yields the smallest drop in performance after transfering to a target language compared to the baselines.}
  \label{fig:gap_all}
\end{figure*}

Where \(P_{TL}\) and \(P_{SL}\) represent the evaluation performance of the exact same method on the target and source-language test sets, respectively. The performance gap values are shown in Figure \ref{fig:gap_all}. We can observe that for most source-language data budgets, we obtain a reduced average transfer gap \(\bar \Delta \%\) through IC-XLT compared to ZS-XLT, 1S-XLT\(^\nabla\) and IC-XLT\(_{SRC}\). 

\begin{table}
\centering
\resizebox{\columnwidth}{!}{%
\begin{tabular}{l|llll}
\hline
$K_{src}$ & \multicolumn{4}{c}{1S} \\
\hline
0 & \multicolumn{4}{c}{28.42} \\
\hline
 & ZS-XLT & 1S-XLT$^{\nabla}$ & IC-XLT$^{M=10}$ & IC-XLT$^{M=20}$ \\
 \hline
8 & 25.40 & 27.12 & \textbf{33.34} & 16.64 \\
16 & 30.84 & 32.63 & \textbf{48.66} & 47.04 \\
32 & 43.56 & 43.36 & 58.91 & \textbf{61.00} \\
64 & 55.15 & 53.85 & \textbf{65.64} & 65.28 \\
Full & 70.15 & 70.39 & \textbf{75.25} & 75.05 \\
\hline
\end{tabular}
}
\caption{\label{tab:acd_results}
Average \(F_1\) micro across 5 target languages for Aspect Category Detection.
}
\end{table}

\begin{table}
\centering
\resizebox{\columnwidth}{!}{%
\begin{tabular}{l|llll}
\hline
$K_{src}$ & \multicolumn{4}{c}{1S} \\
\hline
0 & \multicolumn{4}{c}{30.48} \\
\hline
 & ZS-XLT & 1S-XLT$^{\nabla}$ & IC-XLT$^{M=10}$ & IC-XLT$^{M=20}$ \\
 \hline
8 & 44.05 & 45.86 & 62.46 & \textbf{66.29} \\
16 & 51.13 & 51.97 & 70.30 & \textbf{73.14} \\
32 & 64.74 & 65.28 & 73.27 & \textbf{76.72} \\
64 & 63.45 & 63.68 & 77.55 & \textbf{79.09} \\
Full & 76.12 & 76.21 & \textbf{81.24} & 80.26 \\
\hline
\end{tabular}
}
\caption{\label{tab:massive_results}
Average \(F_1\) micro across 13 target languages for MASSIVE (Domain Classification). 
}
\end{table}

\paragraph{Improvement due to target-language demonstrations.}

\begin{table*}[]
\centering
\resizebox{2.\columnwidth}{!}{%
\begin{tabular}{ll|l|lllll|llll|llll}
\hline
 &  &  & \multicolumn{5}{c}{<5 B} & \multicolumn{4}{c}{5-100 B} & \multicolumn{4}{c}{> 100B} \\
 \hline
 & Method (\(K_{tgt}\)) & Target avg & AMH & AZE & ISL & SWA & URD &ARA & IND &  TUR & THA & FRA & JAP & RUS & SPA \\
 \hline
\multirow{2}{*}{\(\delta \%\)} & 8S-XLT\(^\nabla\) (8) & 0.87 & 4.56 & 0.7 & 1.3 & 1.57 & 2.24 & 0.66 & -0.18 & 0.41 & 0.73 & 0.34 & -0.17 & 0.23 & 0.24 \\
 & IC-XLT (1) & \textbf{9.81} & \textbf{11.5} & \textbf{15.41} & \textbf{17.12} & \textbf{39.55} & \textbf{13.15} & \textbf{7.9} & \textbf{5.63} & \textbf{9.11} & \textbf{2.48} & \textbf{3.28} & \textbf{2.04} & \textbf{2.1} & \textbf{3.25} \\
 \hline
\end{tabular}
}
\caption{
\(\delta \%\) in MASSIVE. The first row refers to the number of tokens of the target languages in mT5 pretraining corpora. We use IC-XLT with \(M=10\).
}
\label{tab:deltas_massive}
\end{table*}

Aiming to assess the proposed method's capacity to utilize demonstrations in the target language to enhance performance in that language, we compute the average percentage improvement in performance \(\delta \%\) after introducing target-language shots. The formula for computing this value is described in Appendix \ref{sec:improvement_due_to_target}.

For the MASSIVE dataset, since 1S-XLT\(^\nabla\) offers minimal improvements with \(K_{tgt}=1\), we conduct a comparison between IC-XLT with \(K_{tgt}=1\) and 8S-XLT\(^\nabla\) with \(K_{tgt}=8\).

Values for \(\delta \%\) are detailed in Tables \ref{tab:deltas_acd} and \ref{tab:deltas_massive} for the Aspect Category Detection (ACD) and MASSIVE datasets, respectively. We observe that IC-XLT consistently achieves the highest (\(\delta \%\)) compared to the fine-tuned approaches. This result underscores IC-XLT's effectiveness for Few-Shot Cross-lingual text classification, a high-level task where methods that involve fine-tuning for language adaptation underperform when the amount of target-language data is limited.

\begin{table}[]
\centering
\resizebox{1.0\columnwidth}{!}{%
\begin{tabular}{ll|l|lllll}
\hline
 & Method (\(K_{tgt}\)) & Target avg & FRA & NLD & RUS & SPA & TUR \\
 \hline
 \multirow{2}{*}{\(\delta \%\)} & 1S-XLT\(^\nabla\) (1) & 0.34 & \textbf{0.03} & 0.46 & -0.22 & 0.31 & 1.10 \\
 & IC-XLT (1) & \textbf{1.92} & 0.01 & \textbf{1.1} & \textbf{0.62} & \textbf{2.04} & \textbf{6.06} \\
 \hline
\end{tabular}
}
\caption{
\(\delta \%\) in the Aspect Category Detection dataset. In here we use IC-XLT with \(M=10\).
}
\label{tab:deltas_acd}
\end{table}

\paragraph{Correlation of methods, pretraining data and linguistic distance.}
\begin{table}[]
\centering
\resizebox{0.8\columnwidth}{!}{%
\begin{tabular}{lll}
\hline
 & mT5 pretraining & Ling. Dist. vs ENG \\
 \hline
IC-XLT$_{SRC}^{M = 10}$ & \textbf{-0.786}$^{**}$ & 0.232 \\
8S-XLT$^{\nabla}$ & \textbf{-0.874}$^{**}$ & 0.331 \\
\hline
\end{tabular}
}
\caption{
Improvement per language correlation (\(\delta \%\)) with Linguistic Distance and language representation in mT5 pretraining corpora (MASSIVE dataset).
}
\label{tab:correlations}
\end{table}

By analyzing results is the MASSIVE dataset across the 13 target languages, we observe that by introducing target-language demonstrations some languages, such as French, Japanese, or Russian, show modest enhancements of around 2\%, while others, like Azeri, Icelandic, and Swahili, benefit from increases exceeding 15\%. To understand the underlying factors contributing to these differences, we explore the relationship between the improvement observed (\(\delta \%\)) and two variables: (1) the number of tokens representing each language in mT5 pretraining corpora \cite{mt5}, and (2) the linguistic distance between each target language and English, the source language, according to the URIEL database \cite{uriel}. The analysis is limited to the MASSIVE dataset because the ACD dataset includes only 5 target languages, mostly European and all well-represented in mT5 pretraining data, with no languages having low representation. This limits the scope for obtaining reliable correlation measures.

To quantitatively assess these relationships, we measure the Spearman Correlation to identify how the token counts of pretraining data and the linguistic proximity to English correlate with the effectiveness of target-language demonstrations in improving cross-lingual transfer performance. 
From the correlations presented in Table \ref{tab:correlations} we can observe that both IC-XLT and 8S-XLT\(^\nabla\) exhibit a statistically significant negative Spearman correlation between the improvement (\(\delta \%\)) and the representation of target languages in the mT5 pretraining corpora. This pattern indicates that languages with less representation in the pretraining data will experience more substantial improvements through target-language adaptation. Correlation with linguistic distance, on the other hand, is weaker and not statistically significant.

\section{Conclusion}
In this paper, we investigated the application of In-Context Tuning for One-Shot Cross-lingual transfer, introducing In-Context Cross-lingual Transfer (IC-XLT). Our evaluations conducted on a multilingual encoder-decoder model (mT5) demonstrate the efficacy of the proposed method in effectively adapting at inference time to target languages using only a One-Shot demonstration in-context, all without incurring additional computational expenses (gradient free). Furthermore, in comparison to ZS-XLT and 1S-XLT\(^\nabla\), IC-XLT demonstrated superior performance and smaller transfer gap for the task of text classification, a  high-level task where FS-XLT tends to underperform.

In scenarios with limited source-language training data, we provide empirical evidence that IC-XLT learns better the source language at the meta-training stage and demonstrates a smaller transfer gap at the adaptation stage with the One-Shot demonstration, compared to ZS-XLT and 1S-XLT\(^\nabla\). This makes IC-XLT a valuable tool for cross-lingual transfer in resource-limited scenarios. Our findings also show a significant correlation between the performance improvements in target languages and their token count in the mT5 pretraining corpus, indicating that languages with lesser representation tend to benefit more from target-language adaptation through IC-XLT. 

To our knowledge, this study represents the first exploration of In-Context Tuning for Cross-Lingual Transfer. For future work, we aim to explore the potential and limitations of this approach by evaluating its applicability to other architectures, such as decoder-only or encoder-only models, and examining the impact of training with a greater number of examples in-context. 

\section{Limitations} 

In this study, we implement our approach using an mT5-large encoder-decoder model. However, an evaluation of its applicability to encoder-only or decoder-only models remains unexplored and it is left for future work. Furthermore, due to storage and compute constraints and the need to conduct experiments across diverse seeds and training data budgets, we opted to fine-tune the models using LoRA \cite{lora}. While some variability compared to the fully trained model is expected with this architectural choice, empirical evidence from \citet{lora} suggests that its impact is minimal.
In this work, we do not compare with methods that translate the source-language training set into target languages. Such approaches require a separate machine translation system and thus are more expensive, falling beyond the scope of our research. Our focus remains on utilizing a single model in an end-to-end manner.
Finally, it is important to outline that due to the maximum input length of mT5 (1024 tokens), scaling IC-XLT is to a larger number of target-language shots (e.g \(K_{tgt}\in\{4,8,16\}\)) may prove difficult using the current approach.  This challenge is particularly pronounced in scenarios with a substantial number of labels, where input text may need to be truncated. Consequently, there is a need to devise a strategy to either reduce input length or integrate information from different example batches in order to address this limitation.

\section*{Acknowledgements}
We thank the anonymous reviewers and the meta reviewer at ARR for their valuable feedback. Also, we thank CONAHCYT for the computer resources provided through the INAOE Supercomputing Laboratory’s Deep Learning Platform for Language Technologies and CIMAT Bajio Super-computing Laboratory (\#300832). Sanchez-Vega acknowledges CONAHCYT for its support through the program “Investigadoras e Investigadores por México” (Project ID.11989, No.1311). Villa-Cueva (CVU 1019520) thanks CONAHCYT for the support through the master's degree scholarship at CIMAT.




\bibliography{anthology,custom}

\appendix

\section{Appendix}
\label{sec:appendix}
\subsection{Shot selection}\label{sec:appendix_shotselection}

Similar to \citet{zhao}, with "\(K\)-shot" we refer to selecting \(K\) examples for each of the \(N\) classes. The examples are randomly sampled from the training splits of the datasets. Note that the number of shots per label may not precisely be \(K\) due to underrepresented classes in the training set. This holds true for the ACD dataset, where certain classes may have insufficient samples to meet the per-class \(K\) value. In such cases, the total number of shots per i-th class is determined as \(\min{(K,N_i)}\), where \(N_i\) represents the total number of samples for the i-th class in the dataset. 

Furthermore, since the ACD task involves a multi-label dataset, multi-label examples may add to more than one of the \(N\) buckets simultaneously. Hence, the total number of examples in a \(K\)-shot dataset is in the range \(\left[K,K\times N\right]\).

\subsection{Hyperparameter selection}\label{sec:hpp_selection}

Here, we outline the hyperparameters utilized for fine-tuning models across the two stages of our pipeline. Initially, we detail the hyperparameters specific to fine-tuning models in the source language for both PFT and ICT methods. Subsequently, we address those employed for the gradient-based XLT methods. The LoRA \cite{lora} parameters are \(r=16, \alpha = 32,\) with dropout of 10\%. For all cases, we employ an AdamW optimizer \cite{adamw} with a linear scheduler and a batch size of 8.

\subsubsection{Fine-tuning on source data}\label{sec:hpp_selection_ft}

For the fine-tuning process on both datasets, we explored learning rates within the range of \(lr \in \{3,4,5,6,7,8,10\}\times 10^{-4}\), selecting \(4\times10^{-4}\) for as it performed adequately on both datasets in the source language.

Regarding the number of epochs for training on the full datasets: for the MASSIVE dataset, we fine-tuned models for 10 epochs under both prompt-based Fine-Tuning (PFT) and In-Context Tuning (ICT) training schemes. For the Aspect Category Detection (ACD) dataset, which is considerably smaller, we extended the training duration to 15 epochs for ICT and 25 epochs for PFT. The decision to train the PFT models for more epochs in the ACD dataset was taken because it underperformed when only 15 epochs were used.

For models trained with a reduced quantity of source-language data, we standardized the training process by setting the learning rate to \(5\times10^{-4}\) for all models across both datasets. Given the similar volume of data in these constrained scenarios, we extended the training period to 35 epochs for both datasets to expose the models to sufficient training data.

\subsubsection{Fine-tuned XLT baselines hyperparameter selection.}

For the fine-tuned XLT baselines, we use the model checkpoint fine tuned in the source language, used for ZS-XLT. Training focuses only on the target language, incorporating only the specified number of target-language shots \(K_{tgt}\), and for 1S-XLT\(_{macro}^\nabla\), an equal amount of source-language shots is added.

We evaluated the following learning rates \(lr \in \{0.5,1,5\}\times 10^{-5}\). We observed that the limited target-language examples often led to overfitting and reduced performance (notably when \(K_{tgt}=1\)) with a large number of epochs, especially in some languages such as Russian. For the reported results, the selected learning rates and training durations are as follows:
\begin{itemize}
    \item For both 1S-XLT\(^{\nabla}\) and 1S-XLT\(_{macro}^\nabla\), a learning rate of \(5\times10^{-5}\) is used, training for 1 epoch for models trained with the full source-language dataset and 5 epochs for those trained with limited source-language data.
    \item  For the 8S-XLT\(^{\nabla}\) baseline, given the increased number of examples, we opt for a learning rate of \(1\times10^{-5}\) across 10 epochs.
\end{itemize}

All adaptations used a batch size of 8 and a constant scheduler.

\subsection{Dataset description}\label{sec:appendix_datasets}

In this Section we provide descriptive information of the employed datasets. MASSIVE features parallel language splits, each comprising 11.5k samples in the training partition and 2.97k in the test partition.

However, for the Aspect Category Detection dataset, which is non-parallel, the sample counts vary across languages. Detailed information on these counts is presented in Table \ref{tab:acd_partition_count}.

\begin{table}[]
\centering
\resizebox{0.4\columnwidth}{!}{%
\begin{tabular}{lll}
\hline
 & Train & Test \\
 \hline
English & 2000 & 676 \\
Spanish & 2070 & 881 \\
French & 1664 & 668 \\
Turkish & 1232 & 144 \\
Russian & 3655 & 1209 \\
Dutch & 1722 & 575 \\
\hline
\end{tabular}
}
\caption{Length of the training and test partitions in the Aspect Category Detection Dataset.}
\label{tab:acd_partition_count}
\end{table}

\subsection{Computing the improvement due to target-language demonstrations} \label{sec:improvement_due_to_target}

By computing \(\delta \%\), we aim to measure the average improvement on performance of the evaluated methods after introducing target-language demonstrations. For this we compute the ratio between the model with target-language demonstrations and the Zero-Shot approach. This value is computed using the following formula:
\[ \delta \% = 100 \times \left[P^{few-shot}_{TL} / P^{zero-zhot}_{TL} - 1\right]\]
Where \(P^{few-shot}_{TL}\) and \(P^{zero-shot}_{TL}\) represent the average evaluation performance of the model under the same training scheme (ICT or PFT), with and without the target-language (\(TL\)) examples, respectively. Hence, for prompt-based fine-tuning \(P^{few-shot}_{TL}\) is 1S/8S-XLT\(^\nabla\) and \(P^{zero-shot}_{TL}\) is ZS-XLT, while for In-Context Tuning \(P^{few-shot}_{TL}\) is IC-XLT and \(P^{zero-shot}_{TL}\) is IC-XLT\(_{SRC}\). 


\subsection{Performance metrics per language of the evaluated method on different source language budgets.}\label{sec:appx_allexps}

This section presents the comprehensive results of our evaluations across different target languages and different source data availability settings of ZS-XLT, 1S-XLT\(^{\nabla}\), and IC-XLT, with English serving as the source language. Detailed performance metrics for cross-lingual transfer on the Aspect Category Detection (ACD) dataset are depicted in Table \ref{tab:all_metrics_acd} and results for the MASSIVE dataset are provided in Tables \ref{tab:all_metrics_massive1} and \ref{tab:all_metrics_massive2}.

\begin{figure*}
    \centering
    \includegraphics[width=\linewidth]{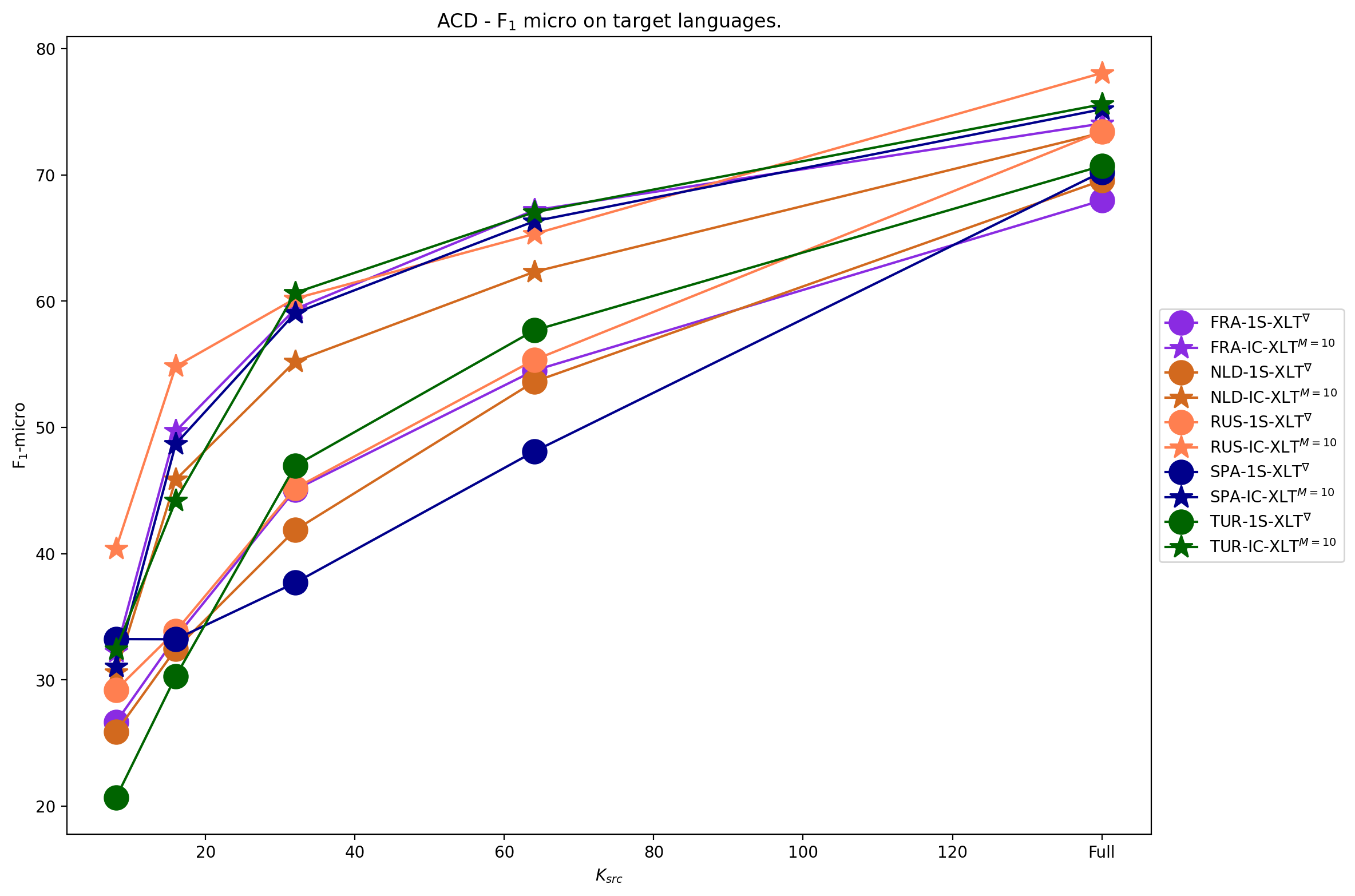}
    \caption{F\(_1\)-micro for each of the 5 evaluated languages in the Aspect Category Detection dataset at different source-language data budget.}
    \label{fig:acd_allplots}
\end{figure*}

\begin{figure*}
    \centering
    \includegraphics[width=\linewidth]{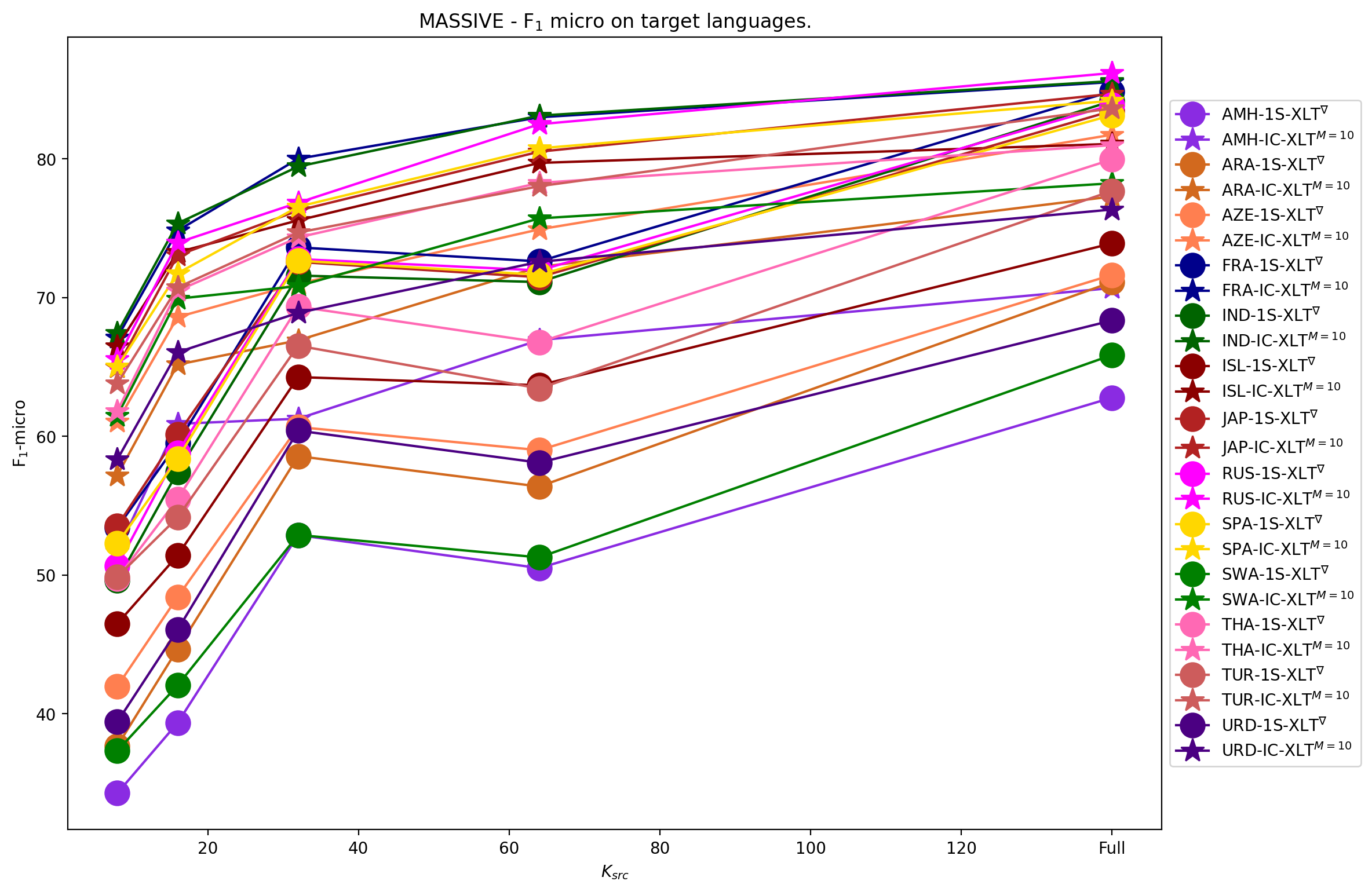}
    \caption{F\(_1\)-micro for each of the 13 evaluated languages in the MASSIVE dataset at different source-language data budget.}
    \label{fig:massive_allplots}
\end{figure*}

\begin{figure*}
\centering
  \begin{subfigure}[b]{0.45\textwidth}
    \includegraphics[width=\linewidth]{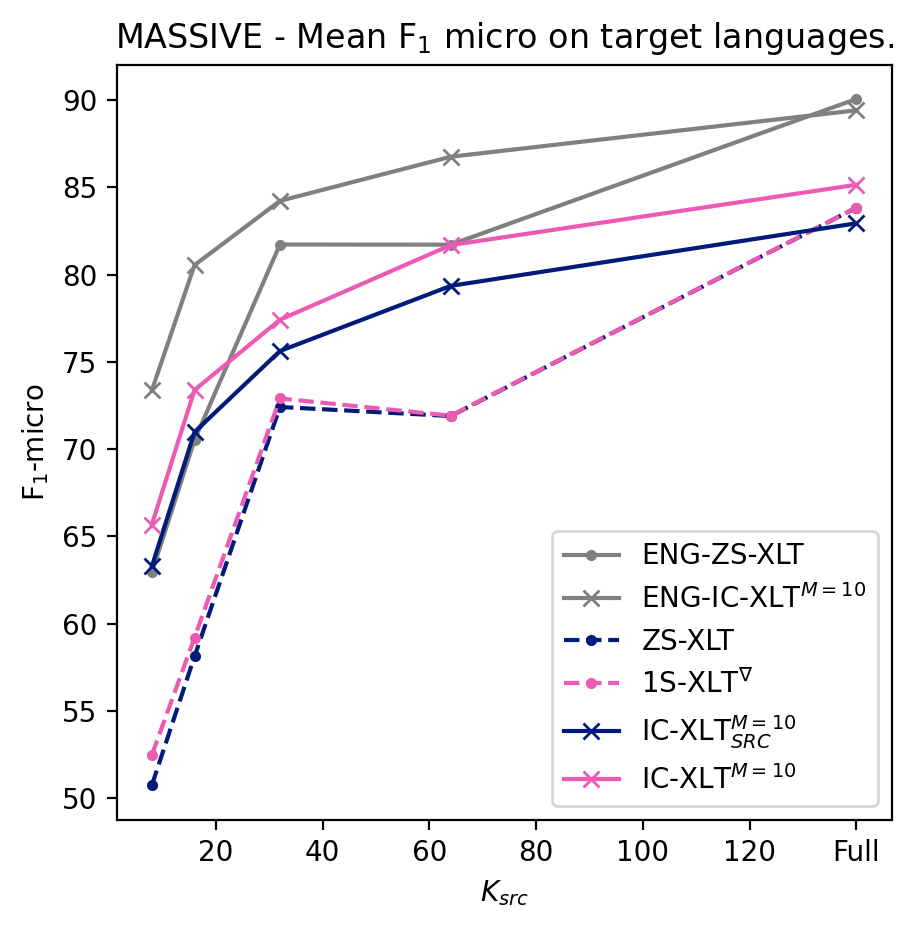}
    \caption{Mean F\(_1\)-micro in the  the MASSIVE dataset across the target languages with large representation in mT5 pretraining corpora (\(>100B\)). In this scenario, the gap between One-Shot (pink) and Zero-Shot (blue) lines is similar as the one observed for ACD in the Figure \ref{fig:acd_plot}.}
    \label{fig:massive_hrl}
  \end{subfigure}
  \quad
    \begin{subfigure}[b]{0.45\textwidth}
    \centering
    \includegraphics[width=\linewidth]{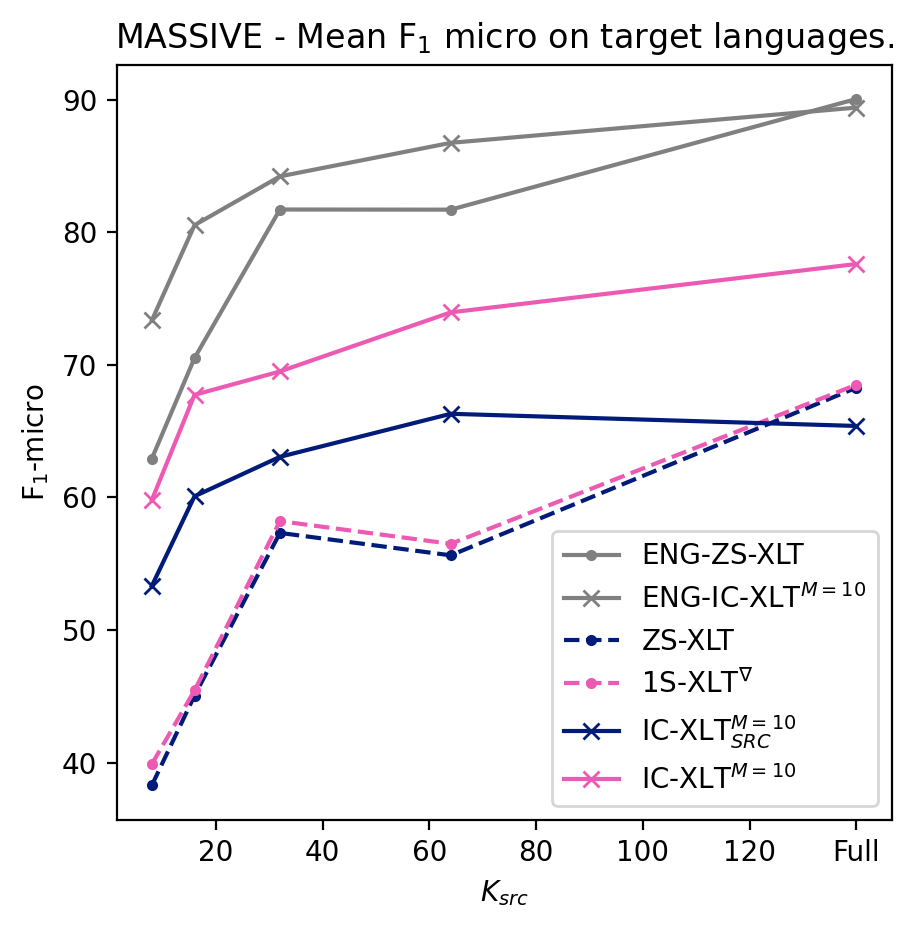}
    \caption{Mean F\(_1\)-micro in the  the MASSIVE dataset across the target languages with smaller representation in mT5 pretraining corpora (\(<100 B\)). We can observe a larger improvement when introducing target-language demonstrations than the one observed for languages well-represented in pretraining (left).}
    \label{fig:massive_lrl}
  \end{subfigure}

  \caption{Comparison of IC-XLT and 1S-XLT\(^{\nabla}\) performance on the MASSIVE datasets at different source-language data budgets on languages with high representation (left), and with low representation (right) in the base model pretraining.}
  \label{fig:both_plots}
\end{figure*}

\begin{figure*}
    \centering  \includegraphics[width=\linewidth]{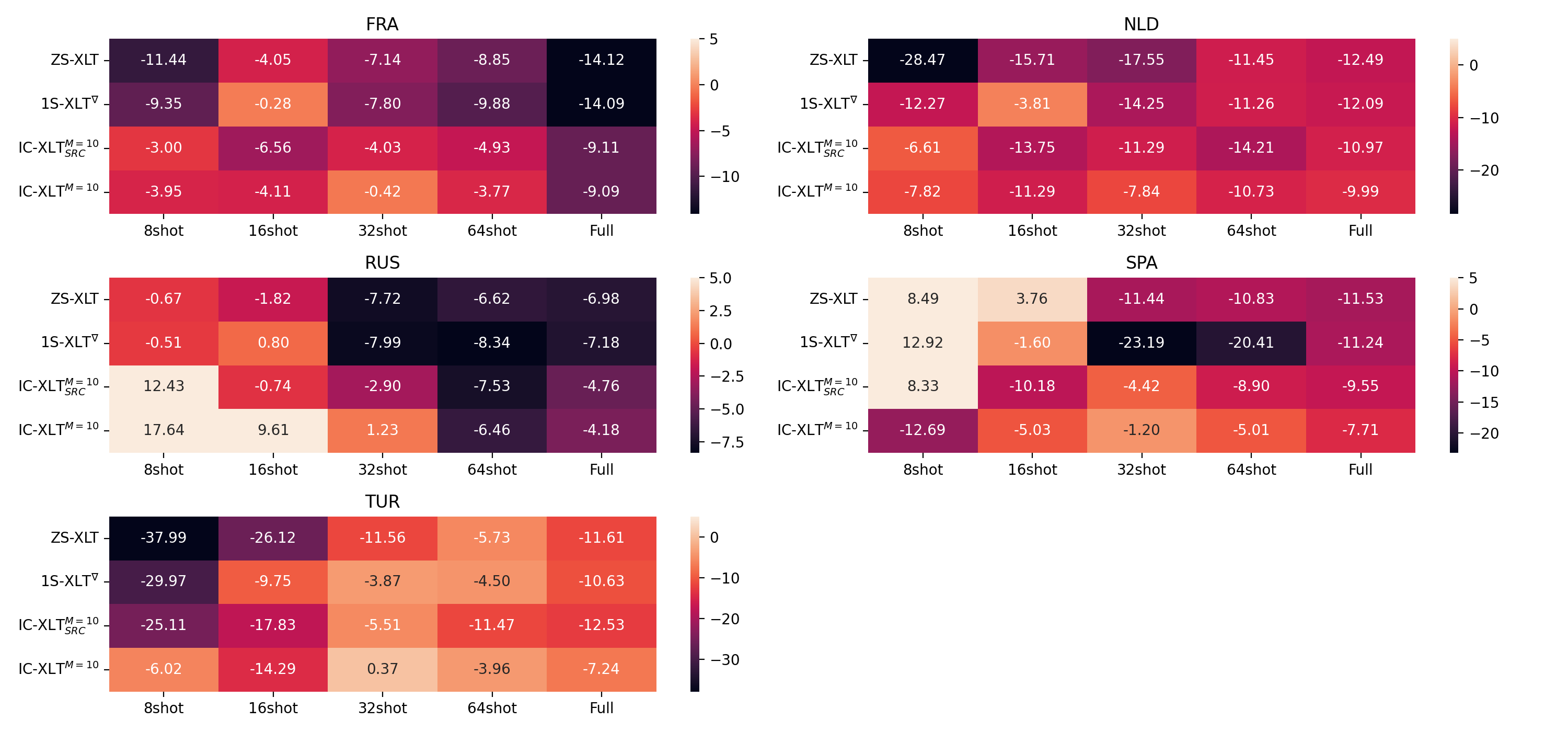}
    \caption{Average Transfer Gap \(\bar \Delta \%\) per language on the Aspect Category Detection dataset.}
    \label{fig:transfers_acd_all}
\end{figure*}

\begin{figure*}
    \centering  \includegraphics[width=\linewidth]{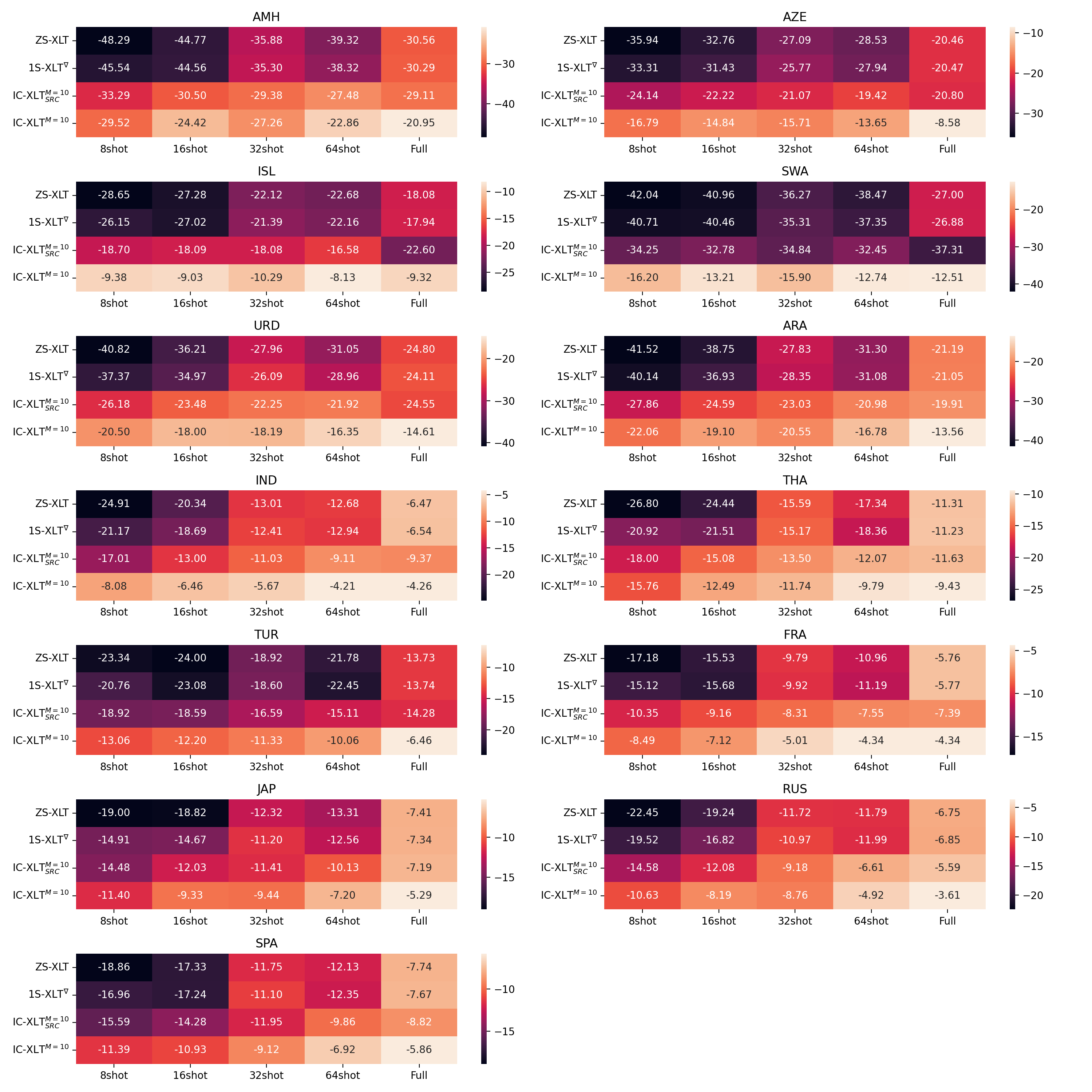}
    \caption{Average Transfer Gap \(\bar \Delta \%\) per language on the MASSIVE dataset.}
    \label{fig:transfers_massive_all}
\end{figure*}

Additionally, we illustrate the language-wise performance at different $K_{src}$ values on Figures \ref{fig:acd_allplots} and \ref{fig:massive_allplots}. Furthermore, for the MASSIVE dataset we illustrate the different behavior of groups of languages with different representation in the mT5 pretraining corpus. Figures \ref{fig:massive_hrl} and \ref{fig:massive_lrl} illustrate the performance metrics for languages with high representation (over 100 billion tokens) and low representation (under 100 billion tokens) in the mT5 pretraining data, respectively. 
We also include the Average Transfer Gaps \(\bar \Delta \%\) per language at the different source-language budgets in Figures \ref{fig:transfers_acd_all} and \ref{fig:transfers_massive_all}.


\begin{table*}[]
\begin{tabular}{lllllll}
\hline
\(K_{src}\) & ENG & FRA & NLD & RUS & SPA & TUR \\
\hline
 & \multicolumn{6}{c}{ZS-XLT (\(K_{tgt} = 0\))} \\
 \hline
8\textit{-shot} & \(29.49 \pm_{2.06}\) & \(26.1 \pm_{1.46}\) & \(21.19 \pm_{3.74}\) & \(29.32 \pm_{2.84}\) & \(31.99 \pm_{2.46}\) & \(18.4 \pm_{4.34}\) \\
16\textit{-shot} & \(33.73 \pm_{4.12}\) & \(32.24 \pm_{3.63}\) & \(28.73 \pm_{6.76}\) & \(33.02 \pm_{3.22}\) & \(34.96 \pm_{4.48}\) & \(25.27 \pm_{7.68}\) \\
32\textit{-shot} & \(49.05 \pm_{5.33}\) & \(45.41 \pm_{4.68}\) & \(40.44 \pm_{4.68}\) & \(45.31 \pm_{5.9}\) & \(43.37 \pm_{4.73}\) & \(43.28 \pm_{3.93}\) \\
64\textit{-shot} & \(60.45 \pm_{3.3}\) & \(55.07 \pm_{2.46}\) & \(53.49 \pm_{1.88}\) & \(56.35 \pm_{1.23}\) & \(53.9 \pm_{3.4}\) & \(56.96 \pm_{3.02}\) \\
Full& \(79.14 \pm_{0.91}\) & \(67.96 \pm_{0.77}\) & \(69.24 \pm_{1.24}\) & \(73.6 \pm_{1.61}\) & \(70.01 \pm_{0.44}\) & \(69.95 \pm_{1.13}\) \\
 \hline
 & \multicolumn{6}{c}{1S-XLT\(^\nabla\) (\(K_{tgt} = 1\))} \\
 \hline
8\textit{-shot} & \(28.97 \pm_{1.66}\) & \(26.65 \pm_{1.09}\) & \(25.87 \pm_{1.56}\) & \(29.21 \pm_{0.56}\) & \(33.23 \pm_{1.33}\) & \(20.66 \pm_{1.42}\) \\
16\textit{-shot} & \(37.62 \pm_{4.98}\) & \(33.37 \pm_{1.98}\) & \(32.44 \pm_{4.32}\) & \(33.86 \pm_{2.48}\) & \(33.22 \pm_{4.23}\) & \(30.28 \pm_{2.66}\) \\
32\textit{-shot} & \(51.94 \pm_{4.06}\) & \(45.07 \pm_{4.69}\) & \(41.88 \pm_{2.26}\) & \(45.18 \pm_{5.88}\) & \(37.69 \pm_{4.04}\) & \(46.97 \pm_{3.1}\) \\
64\textit{-shot} & \(62.32 \pm_{2.53}\) & \(54.5 \pm_{3.37}\) & \(53.62 \pm_{2.33}\) & \(55.34 \pm_{1.48}\) & \(48.09 \pm_{3.17}\) & \(57.68 \pm_{2.65}\) \\
Full& \(79.16 \pm_{0.78}\) & \(67.98 \pm_{0.74}\) & \(69.56 \pm_{0.63}\) & \(73.44 \pm_{1.55}\) & \(70.23 \pm_{0.47}\) & \(70.72 \pm_{1.43}\) \\
 \hline
 & \multicolumn{6}{c}{IC-XLT$_{SRC}^{M=20}$ (\(K_{tgt} = 0\))} \\
 \hline
8\textit{-shot} & \(23.66 \pm_{6.16}\) & \(17.3 \pm_{1.76}\) & \(19.9 \pm_{2.78}\) & \(19.05 \pm_{4.89}\) & \(22.01 \pm_{4.31}\) & \(16.09 \pm_{5.24}\) \\
16\textit{-shot} & \(41.19 \pm_{9.22}\) & \(44.39 \pm_{5.39}\) & \(38.9 \pm_{7.11}\) & \(43.25 \pm_{6.66}\) & \(40.82 \pm_{8.55}\) & \(36.22 \pm_{13.0}\) \\
32\textit{-shot} & \(63.25 \pm_{2.36}\) & \(60.92 \pm_{1.71}\) & \(55.11 \pm_{2.28}\) & \(59.7 \pm_{2.02}\) & \(59.69 \pm_{1.32}\) & \(56.42 \pm_{2.91}\) \\
64\textit{-shot} & \(70.01 \pm_{1.77}\) & \(65.67 \pm_{1.3}\) & \(60.08 \pm_{2.27}\) & \(64.09 \pm_{2.19}\) & \(63.55 \pm_{0.34}\) & \(62.23 \pm_{4.69}\) \\
Full& \(81.76 \pm_{0.81}\) & \(73.49 \pm_{0.37}\) & \(71.89 \pm_{0.38}\) & \(77.52 \pm_{0.38}\) & \(73.14 \pm_{0.82}\) & \(69.51 \pm_{1.04}\) \\
 \hline
 & \multicolumn{6}{c}{IC-XLT$^{M=20}$ (\(K_{tgt} = 1\))} \\
 \hline
8\textit{-shot} & \(23.66 \pm_{6.16}\) & \(17.12 \pm_{13.05}\) & \(15.17 \pm_{6.96}\) & \(16.58 \pm_{11.33}\) & \(13.49 \pm_{8.87}\) & \(20.83 \pm_{13.2}\) \\
16\textit{-shot} & \(41.19 \pm_{9.22}\) & \(48.24 \pm_{4.75}\) & \(44.89 \pm_{6.13}\) & \(49.64 \pm_{4.95}\) & \(47.26 \pm_{8.59}\) & \(45.17 \pm_{8.6}\) \\
32\textit{-shot} & \(63.25 \pm_{2.36}\) & \(61.37 \pm_{2.49}\) & \(58.01 \pm_{1.44}\) & \(62.05 \pm_{1.53}\) & \(61.74 \pm_{1.3}\) & \(61.85 \pm_{5.97}\) \\
64\textit{-shot} & \(70.01 \pm_{1.77}\) & \(65.86 \pm_{1.28}\) & \(63.1 \pm_{0.79}\) & \(65.12 \pm_{0.96}\) & \(65.34 \pm_{1.18}\) & \(66.98 \pm_{2.66}\) \\
Full& \(81.76 \pm_{0.81}\) & \(73.5 \pm_{0.61}\) & \(73.0 \pm_{0.59}\) & \(78.01 \pm_{0.41}\) & \(74.46 \pm_{0.48}\) & \(76.26 \pm_{1.02}\) \\
 \hline
 & \multicolumn{6}{c}{IC-XLT$_{SRC}^{M=10}$ (\(K_{tgt} = 0\))} \\
 \hline
8\textit{-shot} & \(34.71 \pm_{7.33}\) & \(34.05 \pm_{9.33}\) & \(32.13 \pm_{6.46}\) & \(38.42 \pm_{5.92}\) & \(37.47 \pm_{7.7}\) & \(25.75 \pm_{6.15}\) \\
16\textit{-shot} & \(52.08 \pm_{11.85}\) & \(48.49 \pm_{10.46}\) & \(45.12 \pm_{11.08}\) & \(50.81 \pm_{8.61}\) & \(46.89 \pm_{11.15}\) & \(44.34 \pm_{16.01}\) \\
32\textit{-shot} & \(60.45 \pm_{8.84}\) & \(57.6 \pm_{5.95}\) & \(53.42 \pm_{6.53}\) & \(58.15 \pm_{5.02}\) & \(57.32 \pm_{5.48}\) & \(57.22 \pm_{10.22}\) \\
64\textit{-shot} & \(69.84 \pm_{1.32}\) & \(66.96 \pm_{1.43}\) & \(60.41 \pm_{0.79}\) & \(65.14 \pm_{1.08}\) & \(64.17 \pm_{1.5}\) & \(62.31 \pm_{2.33}\) \\
Full& \(81.48 \pm_{0.37}\) & \(74.06 \pm_{1.03}\) & \(72.54 \pm_{0.82}\) & \(77.59 \pm_{0.85}\) & \(73.7 \pm_{0.72}\) & \(71.27 \pm_{1.17}\) \\
 \hline
 & \multicolumn{6}{c}{IC-XLT$^{M=10}$ (\(K_{tgt} = 1\))} \\
 \hline
8\textit{-shot} & \(34.71 \pm_{7.33}\) & \(32.24 \pm_{4.16}\) & \(30.62 \pm_{6.04}\) & \(40.4 \pm_{7.85}\) & \(31.04 \pm_{12.53}\) & \(32.41 \pm_{8.23}\) \\
16\textit{-shot} & \(52.08 \pm_{11.85}\) & \(49.71 \pm_{11.04}\) & \(45.86 \pm_{9.55}\) & \(54.83 \pm_{5.18}\) & \(48.69 \pm_{9.92}\) & \(44.2 \pm_{9.68}\) \\
32\textit{-shot} & \(60.45 \pm_{8.84}\) & \(59.38 \pm_{5.55}\) & \(55.23 \pm_{5.6}\) & \(60.21 \pm_{2.98}\) & \(59.06 \pm_{5.06}\) & \(60.66 \pm_{9.33}\) \\
64\textit{-shot} & \(69.84 \pm_{1.32}\) & \(67.2 \pm_{1.49}\) & \(62.32 \pm_{1.1}\) & \(65.32 \pm_{0.62}\) & \(66.33 \pm_{0.98}\) & \(67.04 \pm_{2.92}\) \\
Full& \(81.48 \pm_{0.37}\) & \(74.07 \pm_{0.55}\) & \(73.34 \pm_{0.82}\) & \(78.07 \pm_{0.76}\) & \(75.2 \pm_{1.33}\) & \(75.59 \pm_{2.84}\) \\
 \hline
\end{tabular}
\caption{Average per language across the different runs for evaluations under different resource budgets for the Aspect Category Detection dataset. In here, $\pm$ refers to the standard deviation of the performance on the conducted runs.}
\label{tab:all_metrics_acd}
\end{table*}

\begin{table*}[]
\centering
\resizebox{2\columnwidth}{!}{%
\begin{tabular}{lllllllll}
\hline
\(K_{src}\) & ENG & AMH & ARA & AZE & FRA & IND & ISL & JAP \\
\hline
 & \multicolumn{8}{c}{ZS-XLT (\(K_{tgt} = 0\))} \\
 \hline
8\textit{-shot} & \(62.93 \pm_{1.5}\) & \(32.53 \pm_{0.42}\) & \(36.81 \pm_{0.93}\) & \(40.3 \pm_{0.82}\) & \(52.11 \pm_{0.77}\) & \(47.24 \pm_{0.64}\) & \(44.9 \pm_{0.99}\) & \(50.96 \pm_{0.67}\) \\
16\textit{-shot} & \(70.52 \pm_{7.24}\) & \(39.27 \pm_{7.79}\) & \(43.38 \pm_{6.55}\) & \(47.55 \pm_{6.48}\) & \(59.71 \pm_{7.75}\) & \(56.42 \pm_{8.62}\) & \(51.42 \pm_{6.8}\) & \(57.27 \pm_{6.1}\) \\
32\textit{-shot} & \(81.72 \pm_{1.39}\) & \(52.4 \pm_{1.73}\) & \(58.98 \pm_{2.25}\) & \(59.58 \pm_{1.63}\) & \(73.72 \pm_{1.88}\) & \(71.08 \pm_{1.46}\) & \(63.67 \pm_{3.83}\) & \(71.65 \pm_{1.33}\) \\
64\textit{-shot} & \(81.71 \pm_{2.81}\) & \(49.68 \pm_{6.32}\) & \(56.19 \pm_{4.85}\) & \(58.49 \pm_{6.41}\) & \(72.78 \pm_{5.1}\) & \(71.34 \pm_{2.6}\) & \(63.23 \pm_{4.88}\) & \(70.85 \pm_{3.34}\) \\
Full& \(90.06 \pm_{0.45}\) & \(62.53 \pm_{0.79}\) & \(70.98 \pm_{1.95}\) & \(71.63 \pm_{1.87}\) & \(84.87 \pm_{0.25}\) & \(84.23 \pm_{0.09}\) & \(73.77 \pm_{0.62}\) & \(83.39 \pm_{0.66}\) \\
\hline
 & \multicolumn{8}{c}{1S-XLT (\(K_{tgt} = 1\))} \\
 \hline
8\textit{-shot} & \(63.3 \pm_{1.4}\) & \(34.26 \pm_{0.62}\) & \(37.67 \pm_{0.9}\) & \(41.96 \pm_{0.3}\) & \(53.41 \pm_{1.02}\) & \(49.6 \pm_{0.57}\) & \(46.48 \pm_{1.34}\) & \(53.53 \pm_{0.32}\) \\
16\textit{-shot} & \(70.52 \pm_{5.74}\) & \(39.33 \pm_{5.89}\) & \(44.65 \pm_{5.74}\) & \(48.4 \pm_{4.75}\) & \(59.49 \pm_{5.58}\) & \(57.42 \pm_{5.93}\) & \(51.4 \pm_{4.01}\) & \(60.13 \pm_{4.87}\) \\
32\textit{-shot} & \(81.7 \pm_{1.06}\) & \(52.87 \pm_{1.07}\) & \(58.56 \pm_{1.65}\) & \(60.66 \pm_{1.92}\) & \(73.61 \pm_{1.31}\) & \(71.58 \pm_{2.2}\) & \(64.26 \pm_{2.96}\) & \(72.56 \pm_{1.24}\) \\
64\textit{-shot} & \(81.17 \pm_{2.61}\) & \(50.49 \pm_{4.91}\) & \(56.37 \pm_{3.92}\) & \(58.99 \pm_{5.67}\) & \(72.6 \pm_{4.53}\) & \(71.11 \pm_{1.85}\) & \(63.67 \pm_{4.49}\) & \(71.46 \pm_{2.95}\) \\
Full& \(90.08 \pm_{0.4}\) & \(62.78 \pm_{0.82}\) & \(71.11 \pm_{1.77}\) & \(71.63 \pm_{1.67}\) & \(84.86 \pm_{0.18}\) & \(84.17 \pm_{0.16}\) & \(73.9 \pm_{0.53}\) & \(83.45 \pm_{0.63}\) \\
\hline
 & \multicolumn{8}{c}{IC-XLT$_{SRC}^{M=20}$ (\(K_{tgt} = 0\))} \\
 \hline
8\textit{-shot} & \(73.24 \pm_{2.71}\) & \(51.95 \pm_{2.84}\) & \(56.6 \pm_{3.29}\) & \(58.92 \pm_{3.18}\) & \(65.75 \pm_{3.04}\) & \(64.79 \pm_{3.74}\) & \(59.41 \pm_{3.12}\) & \(66.87 \pm_{3.61}\) \\
16\textit{-shot} & \(82.0 \pm_{1.37}\) & \(57.2 \pm_{2.99}\) & \(62.97 \pm_{2.75}\) & \(65.03 \pm_{3.15}\) & \(74.38 \pm_{2.1}\) & \(71.95 \pm_{3.03}\) & \(65.89 \pm_{3.72}\) & \(73.25 \pm_{2.3}\) \\
32\textit{-shot} & \(85.03 \pm_{0.52}\) & \(62.5 \pm_{1.5}\) & \(68.45 \pm_{1.87}\) & \(69.26 \pm_{1.61}\) & \(78.52 \pm_{1.14}\) & \(77.93 \pm_{1.89}\) & \(70.62 \pm_{1.53}\) & \(77.94 \pm_{1.11}\) \\
64\textit{-shot} & \(87.18 \pm_{0.66}\) & \(63.94 \pm_{1.44}\) & \(70.06 \pm_{1.4}\) & \(71.18 \pm_{1.89}\) & \(81.37 \pm_{1.14}\) & \(80.31 \pm_{1.44}\) & \(71.16 \pm_{1.45}\) & \(80.6 \pm_{1.19}\) \\
Full& \(89.46 \pm_{0.43}\) & \(61.27 \pm_{2.66}\) & \(70.77 \pm_{2.18}\) & \(71.49 \pm_{2.12}\) & \(83.13 \pm_{1.88}\) & \(81.6 \pm_{2.15}\) & \(68.83 \pm_{3.72}\) & \(82.83 \pm_{2.19}\) \\
\hline
 & \multicolumn{8}{c}{IC-XLT$^{M=20}$ (\(K_{tgt} = 1\))} \\
 \hline
8\textit{-shot} & \(73.24 \pm_{2.71}\) & \(58.17 \pm_{3.39}\) & \(61.45 \pm_{3.41}\) & \(66.0 \pm_{3.14}\) & \(67.26 \pm_{3.72}\) & \(70.46 \pm_{3.35}\) & \(67.16 \pm_{2.72}\) & \(69.77 \pm_{3.16}\) \\
16\textit{-shot} & \(82.0 \pm_{1.37}\) & \(63.5 \pm_{3.67}\) & \(67.55 \pm_{1.93}\) & \(72.78 \pm_{1.86}\) & \(75.98 \pm_{1.83}\) & \(77.77 \pm_{1.79}\) & \(74.27 \pm_{2.1}\) & \(76.95 \pm_{1.29}\) \\
32\textit{-shot} & \(85.03 \pm_{0.52}\) & \(67.12 \pm_{2.39}\) & \(72.76 \pm_{1.33}\) & \(75.42 \pm_{1.79}\) & \(80.06 \pm_{1.06}\) & \(81.6 \pm_{1.21}\) & \(77.81 \pm_{1.06}\) & \(80.13 \pm_{0.86}\) \\
64\textit{-shot} & \(87.18 \pm_{0.66}\) & \(70.14 \pm_{1.6}\) & \(74.69 \pm_{0.98}\) & \(78.02 \pm_{1.65}\) & \(83.29 \pm_{0.79}\) & \(83.9 \pm_{0.84}\) & \(79.74 \pm_{1.08}\) & \(82.09 \pm_{0.7}\) \\
Full& \(89.46 \pm_{0.43}\) & \(68.09 \pm_{4.07}\) & \(76.43 \pm_{1.87}\) & \(81.37 \pm_{1.99}\) & \(84.92 \pm_{1.4}\) & \(85.46 \pm_{1.62}\) & \(80.32 \pm_{1.1}\) & \(83.77 \pm_{1.72}\) \\
\hline
 & \multicolumn{8}{c}{IC-XLT$_{SRC}^{M=10}$ (\(K_{tgt} = 0\))} \\
 \hline
8\textit{-shot} & \(73.36 \pm_{0.92}\) & \(48.95 \pm_{1.98}\) & \(52.93 \pm_{1.57}\) & \(55.65 \pm_{1.65}\) & \(65.77 \pm_{1.16}\) & \(60.89 \pm_{2.12}\) & \(59.64 \pm_{0.97}\) & \(62.74 \pm_{1.29}\) \\
16\textit{-shot} & \(80.54 \pm_{0.99}\) & \(55.99 \pm_{3.32}\) & \(60.75 \pm_{3.44}\) & \(62.65 \pm_{2.91}\) & \(73.17 \pm_{2.18}\) & \(70.08 \pm_{2.66}\) & \(65.99 \pm_{2.78}\) & \(70.86 \pm_{2.6}\) \\
32\textit{-shot} & \(84.22 \pm_{0.62}\) & \(59.47 \pm_{1.84}\) & \(64.83 \pm_{1.4}\) & \(66.48 \pm_{1.8}\) & \(77.22 \pm_{0.99}\) & \(74.92 \pm_{1.29}\) & \(68.99 \pm_{1.54}\) & \(74.61 \pm_{1.08}\) \\
64\textit{-shot} & \(86.75 \pm_{0.29}\) & \(62.92 \pm_{1.28}\) & \(68.55 \pm_{1.43}\) & \(69.91 \pm_{1.7}\) & \(80.21 \pm_{0.78}\) & \(78.86 \pm_{1.54}\) & \(72.37 \pm_{1.16}\) & \(77.97 \pm_{1.69}\) \\
Full& \(89.41 \pm_{0.4}\) & \(63.39 \pm_{3.54}\) & \(71.62 \pm_{2.13}\) & \(70.82 \pm_{3.95}\) & \(82.81 \pm_{1.8}\) & \(81.04 \pm_{2.38}\) & \(69.22 \pm_{3.4}\) & \(82.99 \pm_{1.53}\) \\
\hline
 & \multicolumn{8}{c}{IC-XLT$^{M=10}$ (\(K_{tgt} = 1\))} \\
 \hline
8\textit{-shot} & \(73.36 \pm_{0.92}\) & \(51.7 \pm_{1.88}\) & \(57.18 \pm_{2.67}\) & \(61.04 \pm_{2.01}\) & \(67.12 \pm_{1.62}\) & \(67.44 \pm_{2.98}\) & \(66.48 \pm_{1.45}\) & \(64.99 \pm_{1.72}\) \\
16\textit{-shot} & \(80.54 \pm_{0.99}\) & \(60.89 \pm_{3.56}\) & \(65.16 \pm_{2.89}\) & \(68.59 \pm_{2.67}\) & \(74.81 \pm_{1.81}\) & \(75.34 \pm_{1.74}\) & \(73.26 \pm_{1.39}\) & \(73.03 \pm_{2.16}\) \\
32\textit{-shot} & \(84.22 \pm_{0.62}\) & \(61.26 \pm_{1.67}\) & \(66.91 \pm_{1.4}\) & \(70.99 \pm_{0.92}\) & \(80.0 \pm_{0.73}\) & \(79.44 \pm_{1.07}\) & \(75.56 \pm_{1.17}\) & \(76.27 \pm_{0.44}\) \\
64\textit{-shot} & \(86.75 \pm_{0.29}\) & \(66.93 \pm_{1.62}\) & \(72.2 \pm_{1.22}\) & \(74.91 \pm_{0.94}\) & \(82.99 \pm_{0.78}\) & \(83.11 \pm_{1.18}\) & \(79.7 \pm_{0.67}\) & \(80.51 \pm_{0.78}\) \\
Full& \(89.41 \pm_{0.4}\) & \(70.68 \pm_{2.94}\) & \(77.28 \pm_{0.45}\) & \(81.73 \pm_{1.13}\) & \(85.53 \pm_{1.11}\) & \(85.6 \pm_{0.92}\) & \(81.07 \pm_{0.89}\) & \(84.68 \pm_{0.44}\) \\
\hline
\end{tabular}
}
\caption{
Average performance for English, Amharic, Arabic, Azeri, French, Indonesian, Icelandic, and Japanese across the different runs for evaluations under different resource budgets in the MASSIVE Domain Classification Task. In here, $\pm$ refers to the standard deviation of the performance on the conducted runs.
}
\label{tab:all_metrics_massive1}
\end{table*}

\begin{table*}[]
\centering
\resizebox{2\columnwidth}{!}{%
\begin{tabular}{lllllll}
\hline
\(K_{src}\) & RUS & SPA & SWA & THA & TUR & URD \\
\hline
 & \multicolumn{6}{c}{ZS-XLT (\(K_{tgt} = 0\))} \\
 \hline
8\textit{-shot} & \(48.8 \pm_{1.05}\) & \(51.05 \pm_{0.8}\) & \(36.46 \pm_{0.58}\) & \(46.05 \pm_{0.15}\) & \(48.24 \pm_{0.99}\) & \(37.22 \pm_{0.65}\) \\
16\textit{-shot} & \(57.1 \pm_{7.63}\) & \(58.39 \pm_{7.04}\) & \(41.78 \pm_{6.38}\) & \(53.49 \pm_{7.93}\) & \(53.6 \pm_{5.47}\) & \(45.26 \pm_{7.81}\) \\
32\textit{-shot} & \(72.15 \pm_{2.0}\) & \(72.12 \pm_{1.64}\) & \(52.09 \pm_{2.37}\) & \(69.0 \pm_{2.74}\) & \(66.26 \pm_{1.79}\) & \(58.88 \pm_{2.03}\) \\
64\textit{-shot} & \(72.11 \pm_{5.17}\) & \(71.83 \pm_{4.43}\) & \(50.36 \pm_{5.54}\) & \(67.6 \pm_{5.01}\) & \(63.97 \pm_{5.46}\) & \(56.4 \pm_{4.64}\) \\
Full& \(83.98 \pm_{0.49}\) & \(83.09 \pm_{0.79}\) & \(65.75 \pm_{2.19}\) & \(79.88 \pm_{2.37}\) & \(77.7 \pm_{1.19}\) & \(67.72 \pm_{0.96}\) \\
\hline
 & \multicolumn{6}{c}{1S-XLT (\(K_{tgt} = 1\))} \\
 \hline
8\textit{-shot} & \(50.65 \pm_{1.06}\) & \(52.26 \pm_{1.06}\) & \(37.31 \pm_{0.61}\) & \(49.75 \pm_{0.38}\) & \(49.86 \pm_{0.89}\) & \(39.4 \pm_{0.32}\) \\
16\textit{-shot} & \(58.79 \pm_{6.61}\) & \(58.37 \pm_{5.28}\) & \(42.03 \pm_{4.64}\) & \(55.46 \pm_{6.02}\) & \(54.14 \pm_{3.8}\) & \(46.04 \pm_{5.9}\) \\
32\textit{-shot} & \(72.77 \pm_{2.13}\) & \(72.65 \pm_{1.03}\) & \(52.88 \pm_{2.07}\) & \(69.34 \pm_{2.89}\) & \(66.52 \pm_{1.94}\) & \(60.4 \pm_{1.36}\) \\
64\textit{-shot} & \(71.95 \pm_{4.93}\) & \(71.64 \pm_{3.48}\) & \(51.26 \pm_{4.19}\) & \(66.78 \pm_{4.84}\) & \(63.44 \pm_{5.25}\) & \(58.09 \pm_{3.61}\) \\
Full& \(83.89 \pm_{0.42}\) & \(83.15 \pm_{0.58}\) & \(65.85 \pm_{1.84}\) & \(79.95 \pm_{1.93}\) & \(77.68 \pm_{1.09}\) & \(68.34 \pm_{0.74}\) \\
\hline
 & \multicolumn{6}{c}{IC-XLT$_{SRC}^{M=20}$ (\(K_{tgt} = 0\))} \\
 \hline
8\textit{-shot} & \(64.98 \pm_{3.46}\) & \(63.79 \pm_{3.12}\) & \(51.51 \pm_{4.17}\) & \(63.69 \pm_{3.56}\) & \(62.52 \pm_{3.31}\) & \(57.68 \pm_{2.64}\) \\
16\textit{-shot} & \(74.26 \pm_{2.59}\) & \(72.2 \pm_{2.17}\) & \(55.0 \pm_{4.2}\) & \(69.61 \pm_{2.29}\) & \(68.52 \pm_{2.35}\) & \(63.29 \pm_{2.7}\) \\
32\textit{-shot} & \(79.78 \pm_{0.82}\) & \(75.58 \pm_{1.77}\) & \(59.21 \pm_{3.66}\) & \(75.42 \pm_{1.86}\) & \(73.58 \pm_{1.65}\) & \(68.19 \pm_{1.2}\) \\
64\textit{-shot} & \(81.58 \pm_{1.05}\) & \(78.44 \pm_{1.66}\) & \(60.89 \pm_{2.89}\) & \(77.9 \pm_{1.3}\) & \(76.14 \pm_{1.75}\) & \(70.03 \pm_{0.95}\) \\
Full& \(83.76 \pm_{1.68}\) & \(81.23 \pm_{2.55}\) & \(53.76 \pm_{5.18}\) & \(78.91 \pm_{1.62}\) & \(76.36 \pm_{2.95}\) & \(68.26 \pm_{2.45}\) \\
\hline
 & \multicolumn{6}{c}{IC-XLT$^{M=20}$ (\(K_{tgt} = 1\))} \\
 \hline
8\textit{-shot} & \(69.41 \pm_{3.01}\) & \(67.04 \pm_{3.46}\) & \(65.27 \pm_{3.1}\) & \(66.53 \pm_{2.65}\) & \(70.03 \pm_{3.31}\) & \(63.22 \pm_{2.82}\) \\
16\textit{-shot} & \(77.11 \pm_{1.51}\) & \(75.4 \pm_{1.5}\) & \(71.36 \pm_{2.28}\) & \(72.55 \pm_{0.81}\) & \(76.18 \pm_{1.43}\) & \(69.46 \pm_{1.98}\) \\
32\textit{-shot} & \(81.43 \pm_{1.16}\) & \(78.68 \pm_{1.2}\) & \(74.61 \pm_{1.26}\) & \(76.1 \pm_{2.19}\) & \(78.46 \pm_{1.28}\) & \(73.13 \pm_{1.19}\) \\
64\textit{-shot} & \(83.24 \pm_{0.8}\) & \(81.06 \pm_{1.33}\) & \(76.78 \pm_{1.47}\) & \(79.36 \pm_{0.97}\) & \(80.75 \pm_{1.52}\) & \(75.13 \pm_{0.97}\) \\
Full& \(84.39 \pm_{2.2}\) & \(83.65 \pm_{1.84}\) & \(75.54 \pm_{5.14}\) & \(81.02 \pm_{1.67}\) & \(82.25 \pm_{2.51}\) & \(76.12 \pm_{1.38}\) \\
\hline
 & \multicolumn{6}{c}{IC-XLT$_{SRC}^{M=10}$ (\(K_{tgt} = 0\))} \\
 \hline
8\textit{-shot} & \(62.66 \pm_{1.71}\) & \(61.93 \pm_{1.47}\) & \(48.24 \pm_{3.18}\) & \(60.16 \pm_{1.59}\) & \(59.48 \pm_{1.44}\) & \(54.15 \pm_{1.66}\) \\
16\textit{-shot} & \(70.82 \pm_{3.03}\) & \(69.05 \pm_{2.3}\) & \(54.17 \pm_{4.41}\) & \(68.4 \pm_{2.33}\) & \(65.58 \pm_{2.76}\) & \(61.64 \pm_{2.6}\) \\
32\textit{-shot} & \(76.48 \pm_{0.93}\) & \(74.15 \pm_{1.1}\) & \(54.89 \pm_{3.2}\) & \(72.85 \pm_{1.34}\) & \(70.25 \pm_{1.89}\) & \(65.48 \pm_{1.62}\) \\
64\textit{-shot} & \(81.02 \pm_{0.89}\) & \(78.2 \pm_{1.22}\) & \(58.61 \pm_{3.63}\) & \(76.28 \pm_{0.93}\) & \(73.64 \pm_{0.57}\) & \(67.74 \pm_{1.23}\) \\
Full& \(84.41 \pm_{0.93}\) & \(81.53 \pm_{2.17}\) & \(56.06 \pm_{3.33}\) & \(79.02 \pm_{1.86}\) & \(76.65 \pm_{2.32}\) & \(67.47 \pm_{3.35}\) \\
\hline
 & \multicolumn{6}{c}{IC-XLT$^{M=10}$ (\(K_{tgt} = 1\))} \\
 \hline
8\textit{-shot} & \(65.57 \pm_{2.6}\) & \(65.0 \pm_{1.37}\) & \(61.48 \pm_{1.71}\) & \(61.81 \pm_{2.62}\) & \(63.79 \pm_{2.42}\) & \(58.32 \pm_{2.48}\) \\
16\textit{-shot} & \(73.95 \pm_{2.8}\) & \(71.74 \pm_{2.7}\) & \(69.91 \pm_{2.1}\) & \(70.48 \pm_{2.28}\) & \(70.72 \pm_{2.4}\) & \(66.05 \pm_{2.08}\) \\
32\textit{-shot} & \(76.83 \pm_{0.94}\) & \(76.54 \pm_{0.66}\) & \(70.82 \pm_{1.86}\) & \(74.33 \pm_{1.03}\) & \(74.68 \pm_{0.97}\) & \(68.89 \pm_{1.27}\) \\
64\textit{-shot} & \(82.49 \pm_{0.89}\) & \(80.75 \pm_{1.2}\) & \(75.7 \pm_{2.2}\) & \(78.26 \pm_{0.56}\) & \(78.02 \pm_{0.9}\) & \(72.57 \pm_{1.3}\) \\
Full& \(86.18 \pm_{0.39}\) & \(84.18 \pm_{1.54}\) & \(78.23 \pm_{1.87}\) & \(80.98 \pm_{0.48}\) & \(83.63 \pm_{1.1}\) & \(76.34 \pm_{0.61}\)\\
\hline
\end{tabular}
}
\caption{
Average performance for Russian, Spanish, Swahili, Thai, Turkish, and Urdu across the different runs for evaluations under different resource budgets in the MASSIVE Domain Classification Task. In here, $\pm$ refers to the standard deviation of the performance on the conducted runs.
}
\label{tab:all_metrics_massive2}
\end{table*}

\subsection{Ethics Statement}
The proposed method helps to improve downstream cross-lingual performance on languages underrepresented in the multilingual model pretraining. However, we believe that data collection for low-resource languages should continue to be a priority for the NLP research community. It remains very important to better integrate linguistic diversity in multilingual models to avoid technological biases and avoid hindering technological development in societies whose language has a limited number of speakers, or limited funding for developing linguistic resources.

On the other hand, we acknowledge the challenge posed by language variants, including regional dialects, which often suffer from underrepresentation. This may result in multilingual models biased towards variants with a larger digital footprint, excluding linguistic features of communities less present in digital spaces. We wish to acknowledge this aspect and warn that it could increase the exclusion of these communities from integration into the digital content ecosystem and information technology tools.

\subsection{Licences of systems and datasets}
In this work, the tools utilized include an mT5 model and the \textit{transformers} library \cite{wolf_huggingface}, both of which use the Apache 2.0 license. The MASSIVE dataset, on the other hand, operates under a CC by 4.0 license. As for the Aspect Category Detection dataset, it employs a MS-NC-No ReD license, which limits its usage strictly to an academic scope. Since the aim of this work is to evaluate the performance of a proposed cross-lingual system, we adhere to all the licenses of the utilized material.

The research presented in this paper is intended for academic purposes, and therefore, we adhere to the licenses governing all utilized materials.

\end{document}